\documentclass{article}

\usepackage[preprint, nonatbib]{nips_2018}

\usepackage[utf8]{inputenc} 
\usepackage[T1]{fontenc}    
\usepackage{hyperref}       
\usepackage{url}            
\usepackage{booktabs}       
\usepackage{amsfonts}       
\usepackage{nicefrac}       
\usepackage{microtype}      

\usepackage{amsmath}
\usepackage{algorithm}
\usepackage[noend]{algpseudocode}
\usepackage{graphicx}
\usepackage{caption}
\usepackage{placeins}
\usepackage{longtable}
\usepackage{booktabs,caption,fixltx2e}
\usepackage[flushleft]{threeparttable}
\usepackage{enumitem}
\usepackage{subfig}

\setlist[itemize]{leftmargin=20pt}
\newlength\myindent
\title{Continuous Learning in \\Single-Incremental-Task Scenarios}

\author{
  Davide Maltoni {\normalfont \&} Vincenzo Lomonaco\\
  Department of Computer Science and Engineering,\\
  University of Bologna,\\
  Via Zamboni, 33, 40126 Bologna BO, Italy \\
  \texttt{\{davide.maltoni, vincenzo.lomonaco\}@unibo.it} \\
}

\begin{document}
\maketitle

\begin{abstract}
  It was recently shown that architectural, regularization and rehearsal strategies can be used to train deep models sequentially on a number of disjoint tasks without forgetting previously acquired knowledge. However, these strategies are still unsatisfactory if the tasks are not disjoint but constitute a single incremental task (e.g., class-incremental learning). In this paper we point out the differences between multi-task and single-incremental-task scenarios and show that well-known approaches such as LWF, EWC and SI are not ideal for incremental task scenarios. A new approach, denoted as AR1, combining architectural and regularization strategies is then specifically proposed. AR1 overhead (in terms of memory and computation) is very small thus making it suitable for online learning. When tested on CORe50 and iCIFAR-100, AR1 outperformed existing regularization strategies by a good margin.
\end{abstract}

\section{Introduction}

During the past few years we have witnessed a renewed and growing attention to Continuous Learning (CL) \cite{Goodfellow2013}\cite{Grossberg2013}\cite{Parisi2018}. The interest in CL it essentially twofold. From the artificial intelligence perspective CL can be seen as another important step towards th{}e grand goal of creating autonomous agents which can learn continuously and acquire new complex skills and knowledge. From a more practical perspective CL looks particularly appealing because it enables two important properties: adaptability and scalability. 
One of the key hallmarks of CL techniques is the ability to update the models by using only recent data (i.e., without accessing old data). This is often the only practical solution when learning on the edge from high-dimensional streaming or ephemeral data, which would be impossible to keep in memory and process from scratch every time a new piece of information becomes available. Unfortunately, when (deep or shallow) neural networks are trained only on new data, they experience a rapid overriding of their weights with a phenomenon known in literature as \emph{catastrophic forgetting} \cite{McCloskey1989}\cite{Ratcliff1990}\cite{French1999}. Despite a number of strategies have been recently proposed to mitigate its disruptive effects, catastrophic forgetting still constitutes the main obstacle toward effective CL implementations.

\subsection{Continuous Learning Strategies}

The sudden interest in CL and its applications, especially in the context of deep architectures, has led to significant progress and original research directions, yet leaving the research community without a common terminology and clear objectives. Here we propose, in line with \cite{Kemker2018} and \cite{Zenke2017}, a three-way fuzzy categorization of the most common CL strategies:
\begin{itemize}
\item \textbf{Architectural strategies}: specific architectures, layers, activation functions, and/or weight-freezing strategies are used to mitigate forgetting. It includes dual-memory models attempting to imitate hippocampus-cortex duality.
\item \textbf{Regularization strategies}: the loss function is extended with loss terms promoting selective consolidation of the weights which are important to retain past memories. Include basic regularization techniques such as weight sparsification, dropout or early stopping.
\item \textbf{Rehearsal strategies}: past information is periodically replayed to the model to strengthen connections for memories it has already learned. A simple approach is storing part of the previous training data and interleaving them with new patterns for future training. A more challenging approach is pseudo-rehearsal with generative models.
\end{itemize}

\begin{figure}[!htb]
  \centering
  \includegraphics[width=0.7\textwidth]{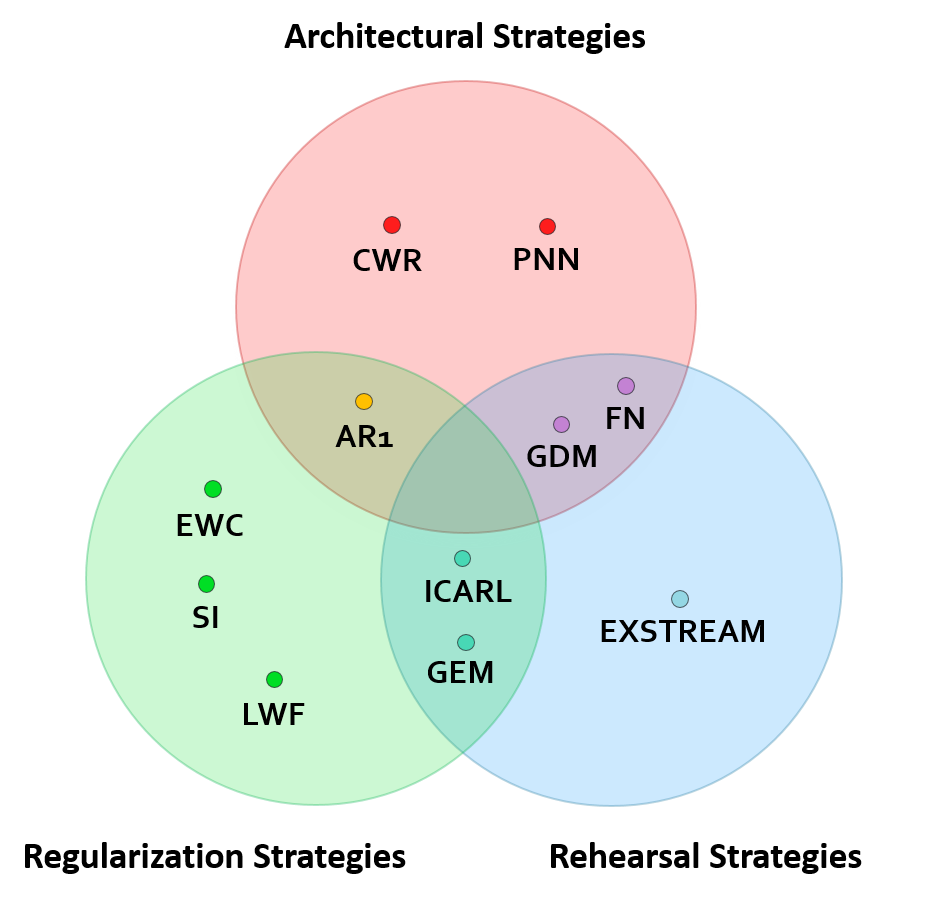}
  \caption{Venn diagram of some of the most popular CL strategies: CWR \cite{pmlr-v78-lomonaco17a}, PNN \cite{Rusu}, EWC \cite{Kirkpatrick2017}, SI \cite{Zenke2017}, LWF \cite{Li2016}, ICARL \cite{Rebuffi2017}, GEM \cite{Lopez-paz2017}, FN \cite{Kemker2018a}, GDM \cite{Parisi2018a}, EXSTREAM \cite{Hayes2018a} and AR1, hereby proposed. Better viewed in color.}
  \label{fig:venn}
\end{figure}

In the Venn diagram of Figure \ref{fig:venn} we show some of the most popular CL strategies. While each category is being populated with an increasing number of novel strategies, there is a large room for yet-to-be-explored techniques especially at the intersection of the three categories.

Progressive Neural Networks (PNN) \cite{Rusu} is one of the first architectural strategies proposed and is based on a clever combination of parameter freezing and network expansion. While PNN was shown to be effective on short series of simple tasks, the number of the model parameters keeps increasing at least linearly with the number of tasks, making it difficult to use for long sequences. The recently proposed CopyWeights with Re-init (CWR) \cite{pmlr-v78-lomonaco17a}, constitutes a simpler and lighter counterpart to PNN (at the cost of a lower flexibility), with a fixed number of shared parameters and already proven to be useful on longer sequences of tasks.

Learning Without Forgetting (LWF) \cite{Li2016} is a regularization strategy attempting to preserve the model accuracy on old tasks by imposing output stability through knowledge distillation \cite{Hinton2015}. Other well-known regularization strategies are Elastic Weights Consolidation (EWC) and Synaptic Intelligence (SI), both articulated around a weighted quadratic regularization loss which penalizes moving weights which are important for old tasks.

At the intersection between rehearsal and regularization strategies we highlight ICARL \cite{Rebuffi2017} and GEM \cite{Lopez-paz2017}. The former, whose acronym stands for \emph{Incremental Classifier and Representation Learning}, includes an external fixed memory to store a subset of old task data based on an elaborated sample selection procedure, but also employs a distillation step acting as a regularization; the latter, known as \emph{Gradient Episodic Memory}, uses a fixed memory to store a subset of old patterns and
add regularization constraints to the loss optimization, aimed not only at controlling forgetting but also at improving accuracy on previous tasks while learning the subsequent ones (a phenomenon known as
``positive backward transfer''). A recent study on memory efficient implementation of pure rehearsal strategies is provided in \cite{Hayes2018a} where a new partitioning-based method for stream clustering named EXSTREAM is shown to be very competitive with a \emph{full rehearsal} approach (storing all the past data) and with other memory management techniques.

Very recently, a growing number of techniques have been proposed on CL based on both variations of the previously introduced strategies or completely novel approaches with different degrees of success (see \cite{Parisi2018} for a review). In particular, FearNet (FN) \cite{Kemker2018a} and Growing Dual-Memory (GDM) \cite{Parisi2018a} are interesting approaches leveraging ideas from architectural and (pseudo) rehearsal categories: a double-memory system is exploited to learn new concepts in a short-term memory and progressively consolidate them in a long-term one.

\subsection{Continuous Learning Benchmarks}

Benchmarking CL strategies today is still highly nonstandard and, even if we focus on supervised classification (e.g. leaving reinforcement learning out), researches often reports their results on different datasets by following different training and evaluation protocols (see Table \ref{tab:mt_vs_sit}). 

Most of CL studies consider a Multi-Task (MT) scenario, where the same model is required to learn incrementally a number of isolated tasks without forgetting how to solve the previous ones. For example, in \cite{Zenke2017} MNIST is split in 5 isolated tasks, where each task consists in learning two classes (i.e. two digits). There is no class overlapping among different tasks, and accuracy is computed separately for each task. Such a model cannot be used to classify an unknown digit into the 10 classes, unless an oracle is available at inference time to associate the unknown pattern to the right task in order to setup the last classification layer(s) accordingly. In other words, these experiments are well suited for studying the feasibility of training a single model on a sequence of disjoint tasks without forgetting how to solve the previous ones, but are not appropriate for addressing incremental problems.

\begin{table}[!htb]
  \caption{Categorizations of CL experiments from the recent literature. Most of the benchmarks are based on reshaped versions of well-known datasets such as MNIST, CIFAR-10, CIFAR-100, ILSVR2012 and CUB-200. CORe50 is one of the few datasets specifically designed for CL in a SIT scenario.}
  \label{tab:mt_vs_sit}
  \centering
  \begin{threeparttable}
  \begin{tabular}{lcc}
    \toprule
    \textbf{Dataset}     & \textbf{Multi-Task (MT)} & \textbf{Single-Incremental-Task (SIT)} \\
    \midrule
    Permuted MNIST 		& \cite{Kirkpatrick2017}\cite{Lopez-paz2017} & -     \\
    Rotated MNIST    	& \cite{Lopez-paz2017} 								& -      \\
    MNIST Split     	& \cite{Zenke2017}										& \cite{Parisi2018}  \\
    CIFAR-10/100 Split  & \cite{Zenke2017} 										& -  \\
    iCIFAR-100     		& \ \ \cite{Lopez-paz2017}${}^\star$							& \cite{Rebuffi2017}  \\
    ILSVRC2012     		& -       													& \cite{Rebuffi2017}  \\
    CUB-200    			& -       													& \cite{Parisi2018}  \\
    CORe50     			& -       													& \cite{pmlr-v78-lomonaco17a} \\
    \bottomrule
  \end{tabular}
    \begin{tablenotes}
      \footnotesize
      \item ${}^\star$In \cite{Lopez-paz2017}, the authors decided to further split the iCIFAR-100 benchmark in 20 different ``tasks'' containing 5 different classes each.
    \end{tablenotes}
    \end{threeparttable}
\end{table}

A still largely unexplored scenario, hereafter denoted as Single-Incremental-Task (SIT) is addressed in \cite{Rebuffi2017} and \cite{pmlr-v78-lomonaco17a}. This scenario considers a single task which is incremental in nature. An example of SIT is the so called class-incremental learning where, we still add new classes sequentially but the classification problem is unique and, when computing accuracy, we need to distinguish among all the classes encountered so far. This is quite common in natural learning, for example in object recognition, as a child learns to recognize new objects, they need to be discriminated w.r.t. the whole set of already known objects (i.e. visual recognition tasks are rarely isolated in nature).

Usually, SIT scenario is more difficult than MT one: in fact, \texttt{i)} we still have to deal with catastrophic forgetting; \texttt{ii)} we need to learn to discriminate classes that typically we never see together (e.g. in the same batch), except when a memory buffer is used to store/replay a fraction of past data.

\begin{figure}[!htb]
  \centering
  \includegraphics[width=\columnwidth]{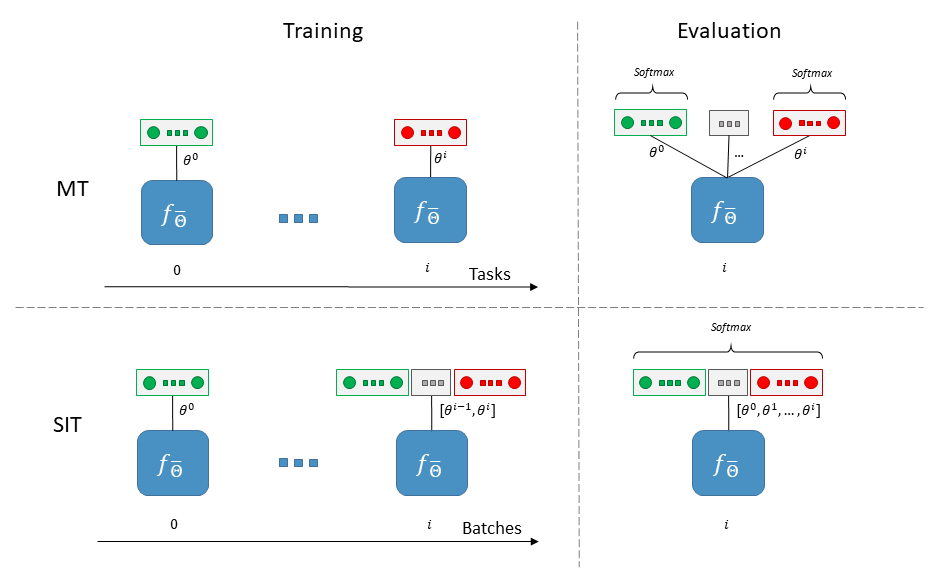}
  \caption{Key architectural differences between MT and SIT scenarios: a disjoint output layer (also denoted as ``head'') is used in MT for each independent task, while a single (dynamically expanded) output layer is used in SIT to include all the classes encountered so far. Better viewed in color.}
  \label{fig:mt_vs_sit}
\end{figure}

Figure \ref{fig:mt_vs_sit} graphically highlights the difference between MT and SIT. While for MT the output neurons are grouped into separate classification layers (one for each task), SIT uses a single output layer including the neurons of all the classes encountered so far. In the MT training phase, the output layer of the  batch can be trained apart while sharing the rest of the model weights (denoted as $ \bar{\Theta}$ in the figure). This is not the case in SIT where weights learned for the old classes could be exploited when learning the current batch classes. In the MT evaluation phase, assuming to know the task membership of each test sample, each task can be assessed separately with the corresponding classification layer. Instead, in the SIT scenario, the evaluation is performed agnostically with respect to the membership of a sample to a specific incremental batch and the final probabilities are computed through a unique softmax layer; this requires to compare objects that were never seen together during training and can have a strong impact on final accuracy. Some researchers, in the continuous learning context, use the term “head” to denote the output classification layer: using this terminology, the MT scenario can be implemented with multiple disjoint heads, while SIT is characterized by a single expanding head.

\begin{figure}[!htb]
  \centering
  \includegraphics[width=\columnwidth]{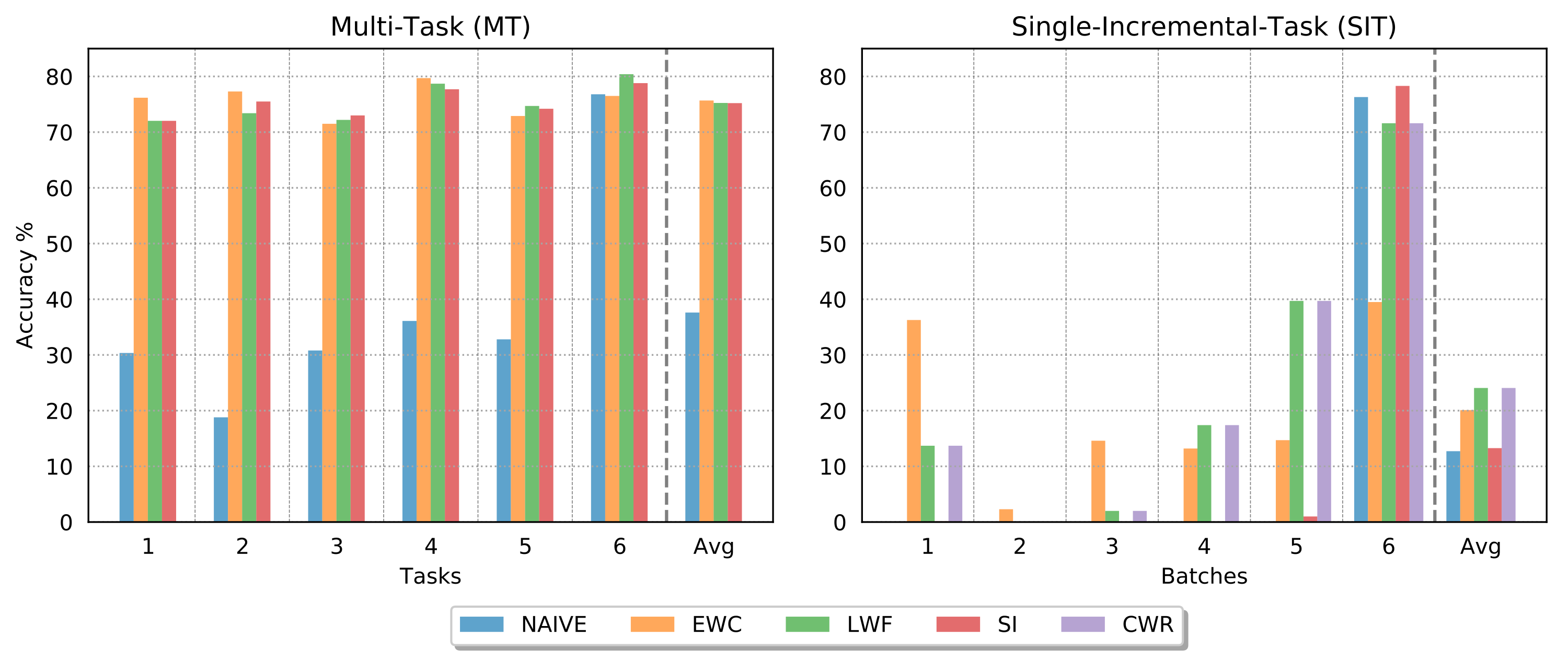}
  \caption{Accuracy results in the MT and SIT scenarios for 5 CL strategies (NAÏVE, EWC, LWF, SI, CWR) after the last training batch. With NAÏVE we denote a simple incremental fine-tuning where early stopping is the only option available to limit forgetting. Analogously to \cite{Zenke2017}, this experiment was performed on the first 6 tasks of CIFAR-10/100 split. For both MT and SIT we report the accuracies on the classes of each batch $(1, 2, \dots, 6)$ and their average (Avg). CWR is specifically designed for SIT and was not tested under MT. Better viewed in color.}
  \label{fig:cifar_split_res}
\end{figure}

Figure \ref{fig:cifar_split_res} provides an example that quantifies how much more complex SIT is than MIT on CIFAR-10/100 Split. For direct comparison with \cite{Zenke2017} here we limit the number of tasks/batches to 6 (instead of 11) and report the accuracy only at the end of training (i.e., after the 6th batch). It is evident that SIT represents a much more difficult challenge for state-of-the-art CL strategies. Looking at the average accuracy we can notice a gap between MT and SIT of more than 30\% regardless the CL technique. Actually, the unbalanced nature of the CIFAR-10/100 Split benchmark (50\% of all the train and test set examples belong to the first batch) makes SIT strategies quite harder to parametrize.  However, as argued by other researches \cite{Kemker2018a}\cite{Kemker2018}, most of the existing CL approaches perform well on MT (with a moderate number of tasks) but fail on complex SIT scenario with several (limited-size) batches.

Finally, it is worth noting that class-incremental learning is not the only setting of interest for SIT strategy; in \cite{pmlr-v78-lomonaco17a} we introduced\footnote{In \cite{pmlr-v78-lomonaco17a} NI, NC and NIC are referred to as scenarios. Since in this paper we use the terms scenario for MT and SIT, to avoid confusion we refer to NI, NC and NIC as specific update content types.} three different update content types:

\begin{itemize}
	\item \textbf{New Instances} (NI): new training patterns of the same classes become available in subsequent batches with new poses and environment conditions (illumination, background, occlusions, etc.).
	\item \textbf{New Classes} (NC): new training patterns belonging to different classes become available in subsequent batches. This coincides with class-incremental learning. 
	\item \textbf{New Instances and Classes} (NIC): new training patterns belonging to both known and new classes become available in subsequent training batches.
\end{itemize}

NI and NIC are often more appropriate than NC in real-world applications; unfortunately, MT approaches cannot operate under these settings.
In the rest of this paper we focus on SIT with NC update content type since it is closer to the experiments reported so far by the research community, leaving other settings for future works. To avoid the cumbersome notation SIT-NC, hereafter we leave the second part implicit.

\subsection{Contribution and Summary}

The main contributions of this paper are here summarized:
\begin{itemize}
	\item We investigate and point out the main differences between MT and SIT scenarios under the NC update content type.
	\item We review some popular CL approaches (LWF, EWC, SI) and the practical issues related to their implementations under SIT scenario.
	\item We propose a new CL strategy (AR1) by first extending the CWR architectural strategy that we introduced in \cite{pmlr-v78-lomonaco17a} and then combining it with a light regularization approach such as SI.
	\item We compare AR1 with other common CL strategies on CORe50 and iCIFAR-100 benchmarks where we exceed by a good margin the accuracy of existing approaches.
	\item We discuss diagnostic techniques to simplify hyperparameters tuning in complex CL settings and to detect misbehaviors.
	\item In the appendices we provide implementation details of all the techniques reviewed/proposed and the experiment carried out using the Caffe framework \cite{Jia2014}.
\end{itemize}

In Section \ref{sec:continuous_learning_strategies_in_sit} we discuss the practical implementation of some popular CL strategies: in particular, we argue that ad-hoc normalization steps are necessary to make them properly work with several sequential batches under SIT scenario. For each technique we also focus on the overhead it requires in terms of computation and storage resources. In Section \ref{sec:ar1_combining_architectural_and_regularization_strategies} we propose a new strategy (denoted as AR1) which combines architectural and regularization techniques and can operate online because of a very low overhead. Section \ref{sec:experiments} and Section \ref{sec:practical_advices_for_hyperparameters_tuning} provide experimental results and compare AR1 with existing techniques on both CORe50 and iCIFAR-100 along with some practical advices for hyperparameters tuning.  Finally, in Section \ref{sec:conclusions} we draw some conclusions and discuss future work.

\section{Continuous Learning Strategies in SIT}
\label{sec:continuous_learning_strategies_in_sit}

In this section we review recent continuous learning techniques, and discuss their implementation in the SIT scenario. In our experience, moving from a few tasks in the MT scenario to many batches in SIT such as with CORe50, requires specific normalization techniques and careful calibration of hyper-parameters. 

In this paper we focus on architectural and regularization strategies\footnote{Actually, in section \ref{subsec:rehe_comp} some preliminary experiments are included where the proposed CL strategy (AR1) is compared with well-known rehearsal techniques such as ICARL and GEM.}: while we recognize the pragmatism and high practical value of rehearsal approaches here we concentrate on learning architectures where past experiences are ``neurally'' encoded; pseudo-rehearsal approaches by generative models is an emerging and very interesting approach but its practical effectiveness still needs to be proved with state-of-the-art deep architectures.

\subsection{Learning Without Forgetting (LWF)}

LWF \cite{Li2016} is a regularization approach which tries to control forgetting by imposing output (i.e. prediction) stability. 

Let us consider an output level with $s$ classes (i.e. $s$ neurons) and assume that some classes were already learnt in previous batches. The current batch $B_i$ includes $n_i$ examples drawn from $s_i$ (still unseen) classes, then LWF:

\begin{itemize}
	\item At the beginning of batch $B_i$, before the training start, computes the prediction of the network for each new pattern in $B_i$. With this purpose it performs a forward pass and stores the $s$-dimensional network prediction $\hat{y}_{lwf}$ for each of the  examples in $B_i$. 
	\item Starts training the network with Stochastic Gradient Descent (SGD) by using a two component loss:
		\begin{equation}
		(1-\lambda) \cdot L_{cross}(\hat{y}, t=\hat{y}_{1h}) + \lambda \cdot L_{kdl}(\hat{y}, t=\hat{y}_{lwf})
		\end{equation}
	where:
		\begin{itemize}
			\item $\hat{y}$ are the network predictions (evolving while the model is trained).
			\item The first part is the usual cross-entropy loss whose target vectors $t$ take the form of one-hot vectors $\hat{y}_{1h}$ corresponding to the true pattern labels. This component adjusts the model weights to learn the new classes in $B_i$.
			\item The second part is a Knowledge Distillation Loss \cite{Hinton2015} which tries to keep the network predictions close to $\hat{y}_{lwf}$ (here used as soft target vectors). This component tries to preserve (for the old classes which are not in the current batch) a stable response. The second term can be replaced with one term for each old task/batch; the two formulations are equivalent but the compact form here proposed is simpler to deal with in practice.
			\item The parameter $\lambda \in [0,1]$ defines the relative weights of the two loss components, thus controlling the trade-off between stability and plasticity.
		\end{itemize}
\end{itemize}

In the MT scenario, $L_{cross}$ is computed only for the $n_i$ new classes in $B_i$, while $L_{kdl}$ is computed for all the ($\sum_{j<i}n_j$) classes previously learned. In SIT there is no such distinction and both the loss components are computed for all the $s$ classes encountered so far.

In our LWF implementation for SIT we:
\begin{itemize}
	\item replaced $L_{kdl}$ with $L_{cross}$ in the second terms. LWF authors argued in \cite{Li2016} that the Knowledge Distillation Loss can be replaced with Cross-Entropy with no significant accuracy change. In our initial experiments we obtained similar results, so for simplicity we adopted cross-entropy.
	\begin{equation}
		L_1 = (1-\lambda) \cdot L_{cross}(\hat{y}, t=\hat{y}_{1h}) + \lambda \cdot L_{cross}(\hat{y}, t=\hat{y}_{lwf})
	\end{equation}
	\item fused the two loss components into a single loss with a weighted soft target vector:
	\begin{equation}
		L_2 = L_{cross}(\hat{y}, t=(1-\lambda) \cdot \hat{y}_{1h} + \lambda \cdot \hat{y}_{lwf})
	\end{equation}
	It can be simply proved that $L_1$ and $L_2$ are equivalent and lead to the same gradient flow. In fact, for cross-entropy the gradient of the loss function with respect to the logit layer $o$ (i.e., the layer before softmax) is $\partial L_{cross}/\partial o = (\hat{y}-t)$ and therefore:
	\begin{align}
		\frac{\partial L_1}{\partial o} = (1 - \lambda) \cdot (\hat{y}-\hat{y}_{1h}) + \lambda \cdot (\hat{y} - \hat{y}_{lwf}) &=\nonumber \\
		(\hat{y} - \hat{y}_{1h}) + \lambda \cdot (\hat{y}_{1h} - \hat{y}_{lwf}) &= \nonumber \\
		\hat{y} - ((1-\lambda) \cdot \hat{y}_{1h} + \lambda \cdot \hat{y}_{lwf}) &= \frac{\partial L_2}{\partial o}.
	\end{align}
	\end{itemize}
	Using a single value of $\lambda$ across the sequential training batches can be suboptimal, since the importance of the past should increase with the number of classes learnt. We experimentally found\footnote{In general the suitability of a weighting scheme, can be assessed by looking at the evolution of confusion matrices (see Figure \ref{fig:cms_lwf} in Section \ref{sec:practical_advices_for_hyperparameters_tuning}). Sophisticated weighted schemes tend to overfit the particular batch sequences and must be adopted with care. Eq. \ref{eq:map} is quite simple and has no hyperparameters to tune.} that a reasonable solution is increasing $\lambda$ according to the proportion of the number of examples in the current batch w.r.t. the number of examples encountered so far. A batch specific value $\lambda_i$ can be obtained as:
	\begin{equation}
		\label{eq:map}
  		\lambda_i =\left\{
  			\begin{array}{@{}ll@{}}
    			0, & i=1 \\
    			map(1-\dfrac{n_i}{\sum_{j\leq i} n_j}), & i>1
  			\end{array}\right.
	\end{equation} 
where $map$ is a linear mapping function that can shift and stretch/compress its input. For example, considering the number of classes in CORe50 (i.e. 10 in the first batch and 5 in the successive batches) and assuming that $map$ is the identity function, we obtain: $\lambda_1=0, \lambda_2=\frac{2}{3}, \lambda_1=\frac{3}{4}, \dots, \lambda_9=\frac{9}{10}$.

Another important facet is the learning strength to adopt in the initial batch $B_1$ and successive batches $B_i$. It is worth noting that in LWF (as for EWC and SI) training on incremental batches $B_i, i>1$ should not be forced to convergence. In fact, as the regularization part of the loss becomes dominant the training accuracy tend to decrease and trying to leverage it with aggressive learning rates and high number of epochs can lead to divergence. In our experiments, we trained the model on each batch for a fixed small number of epochs without forcing convergence. Using a simple early stopping criteria is crucial for continuous learning because of efficiency and lack of realistic validation sets.

Summarizing, LWF implementation with weighted soft target vectors is very simple and, for each batch $B_i, i>1$, its overhead consists of \texttt{i)} computation: one extra forward pass for each of the $n_i$ pattern; \texttt{ii)} storage: temporary storing (for the batch lifespan) the $\hat{y}_{lwf}$ predictions, consisting of $n_i \cdot s$ values.

\subsection{Elastic Weight Consolidation (EWC)}

EWC \cite{Kirkpatrick2017} is a regularization approach which tries to control forgetting by selectively constraining (i.e., freezing to some extent) the model weights which are important for the previous tasks. 

Intuitively, once a model has been trained on a task, thus reaching a minimum in the loss surface, the sensitivity of the model w.r.t. each of its weight $\theta_k$ can be estimated by looking at the curvature of the loss surface along the direction determined by $\theta_k$ changes. In fact, high curvature means that a slight $\theta_k$ change results in a sharp increase of the loss. The diagonal of the Fisher information matrix $F$, which can be computed from first-order derivatives alone, is equivalent to the second derivative (i.e. curvature) of the loss near a minimum. Therefore, the $k^{th}$ diagonal element in $F$ (hereafter denoted as $F_k$) denotes the importance of weight $\theta_k$. Important weights must be moved as little as possible when the model is fine-tuned on new tasks. In a two tasks scenario this can be achieved by adding a regularization term to the loss function when training on the second task: 
\begin{equation}
	L = L_{cross}(\hat{y}, t=\hat{y}_{1h}) + \frac{\lambda}{2} \cdot \sum_k F_k (\theta_k - \theta_k^*)^2
  \label{eq:ewc_loss}
\end{equation}
where:
\begin{itemize}
	\item $\theta_k^*$ are the optimal weight values resulting from the first task.
	\item $\lambda$ is the regularization strength.
\end{itemize}

Let us now consider a sequence of tasks or batches $B_i$. After training the model on batch $B_i$ we need to compute the Fisher information matrix $F^i$ and store the set of optimal weights $\Theta^i$. $F_i$ and $\Theta^i$ will be then used to regularize the training on $B_{i+1}$. Each diagonal element $F_k^i$ can be computed as the variance of $\partial L_{cross}(\hat{y}, t)/\partial \theta_k$ over the $n_i$ patterns of $B_i$. 

Two different EWC implementations can be setup in practice:
\begin{enumerate}
	\item A distinct regulation term is added to the loss function for each old task. This requires maintaining a Fisher matrix $F^i$ and a set of optimal weights $\Theta^i$ for each of the previous tasks/batches;
	\item A single Fisher matrix $F$ is initialized to 0 and consolidated at the end of a batch $B_i$ by (element wise) summing the Fisher information: $F=F+F^i$. A single set of optimal weights $\Theta$ is also maintained by using the most recent ones ($\Theta=\Theta^i$) since $\Theta^i$ already incorporates constraints from all previous batches (refer to the discussion in \cite{Huszar2017}\cite{Kirkpatrick2018}).
\end{enumerate}

Option 1 can be advantageous to precisely control EWC training dynamic in the MT scenario with few tasks, but is not practical (because of storage and computation issues) in SIT scenario with several batches.     
It is worth noting that in option 2, the $F_k$ values can only increase as new batches are processed, potentially leading to divergence for large $\lambda$. To better understand this issue, let us consider how the regularization term is dealt with by gradient descent: this is quite similar to L2 regularization and can be implemented as a special weight decay where weights $\theta_k$ are not decayed toward 0, but toward $\theta_k^*$. The weight update determined by the loss function (eq. \ref{eq:ewc_loss}) is:
\begin{equation}
	\theta_k' = \theta_k - \eta \cdot \dfrac{\partial L_{cross}(\hat{y},t)}{\partial \theta_k} - \eta \cdot F_k (\theta_k - \theta_k^*)
  \label{eq:wheights_update}
\end{equation}
where $\eta$ is the learning rate. In the above equation if, for some $k$, the product $\eta \cdot \lambda \cdot F_k$ is greater than 1, the weight correction toward $\theta_k^*$ is excessive and we overshoot the desired value. Tuning $\lambda$ according to the maximum theoretical value of $F_k$ is problematic because: \texttt{i)} we do not know such value; \texttt{ii)} using a too high value might lead to unsatisfactory performance since does not allow to constrain the weights associated to mid-range $F_k$ enough. We empirically found that a feasible solution is normalizing $F$ after each batch $B_i$ as:
\begin{align}
	F &= F + F^i \nonumber \\ 
	\hat{F} &= clip(\frac{F}{i}, max_F)
\end{align}
where $clip$ sets the matrix values exceeding $max_F$ to the constant $max_F$. Note that $F/i$ replaces the Fisher matrix sum with an average, and this could be counterintuitive. Let us suppose that weight $\theta_5$ is very important for batch $B_1$ and this is reflected by an high value of $F_5^1$, then if $\theta_5$ is not important for $B_2$ as well (i.e., $F_5^2$ is small) computing the average $\smash{1/2 \cdot (F_5^1 + F_5^2)}$ pulls down the combined importance. However, this can be compensated by a proper selection of a $max_F$ in order to saturate $\hat{F}$ values even for those weights which are important for a single task. Given $max_F$ and $\eta$ we can easily determine the maximum value for $\lambda$ as $1/(\eta \cdot max_F)$.

\begin{figure}[!htb]
  \centering
   \includegraphics[width=\textwidth]{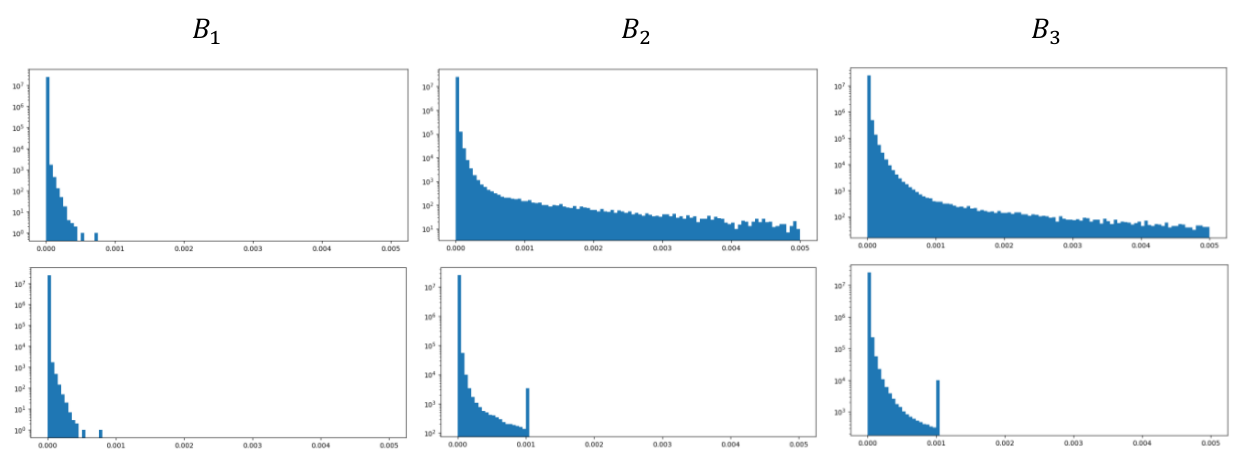}
  \caption{CaffeNet trained by EWC on CORe50 SIT (details in Section \ref{sub:results_on_core50}). The first row shows $F$ values distribution denoting a long tail on the right: considering the logarithmic scale, the number of weights values taking high values in $F$ is quite limited. The second row shows the normalized matrix $\hat{F}$ obtained by averaging $F$ values and max clipping to 0.001. Saturation to 0.001 is well evident, but after $B_3$ the fraction of saturated weights is small (about 1/1000).
  \label{fig:f_values_ewc}}
\end{figure}

An example is shown in Figure \ref{fig:f_values_ewc}, where the distribution of Fisher information values is reported after $B_1$, $B_2$ and $B_3$. In the first row $F$ values denotes a long tail on the right. In the second row, $F_k$ values are averaged and clipped to $0.001$ thus allowing to work with higher $\lambda$ and better control forgetting.

Summarizing, EWC implementation is moderately simple and, for each batch $B_i$, its overhead consists of \texttt{i)} computation of Fisher information $F^i$, requiring one forward and one backward propagation for each of the $n_i$ patterns; \texttt{ii)} storage of $F$ and $\Theta$, totaling $2 \cdot m$ values, where $m$ is the number of model weights (including biases).

\subsection{Synaptic Intelligence}
\label{sub:synaptic_intelligence}

SI was introduced in \cite{Zenke2017} as a variant of EWC. The authors argued that computation of Fisher information is expensive for continuous learning and proposed to calculate weight importance on-line during SGD. 

The loss change given by a single weight update step during SGD is given by:
\begin{equation}
 \Delta L_k = \Delta \theta_k \cdot \frac{\partial L}{\partial\theta_k} 
\end{equation}
where $\Delta\theta_k = \theta_k' - \theta_k$ is the weight update amount and $\partial L / \partial \theta_k$ the gradient. The total loss change associated to a single parameter $\theta_k$ can be obtained as running sum $\sum \Delta L_k$ over the weight trajectory (i.e., the sequence of weight update steps during the training on a batch). 
The weight importance (here denoted as $F_k$ to keep notation uniform with previous section) is then computed as:
\begin{equation}
	F_k = \frac{\sum \Delta L_k}{T_k^2 + \xi}
  \label{eq:Fk}
\end{equation}
where $T_k$ is the total movement of weight $\theta_k$ during the training on a batch (i.e., the difference between its final and the initial value) and $\xi$ is a small constant to avoid division by 0 (see \cite{Zenke2017} for more details). Note that the whole data needed to calculate $F_k$ is available during SGD and no extra computation is needed.

In the SIT scenario we empirically found that an effective normalization after each batch $B_i$ is:
\begin{align}
	F &= F + w_i \cdot F^i \nonumber \\
	\hat{F} &= clip(F, max_F)
  \label{eq:fhat}
\end{align} 
where $F$ is set to 0 before first batch and then consolidated as a weighted sum with batch specific weights $w_i$. Actually in our experiments, as reported in Section \ref{sec:practical_advices_for_hyperparameters_tuning}, we used a small value $w_1$ for the first batch and a constant higher value for all successive batches: $w_2 = w_3 = \dots = w_9$. Considering CORe50 experiments, since in the first batch we tune a model from ImageNet weight initialization, the trajectories that most of the weights have to cover to adapt to CORe50 are longer than for successive batches whose tuning is intra dataset. This is not the case for EWC, because EWC looks at the loss surface at convergence, independently of the length of weight trajectories. 

\begin{figure}[!htb]
  \centering
  \includegraphics[width=\textwidth]{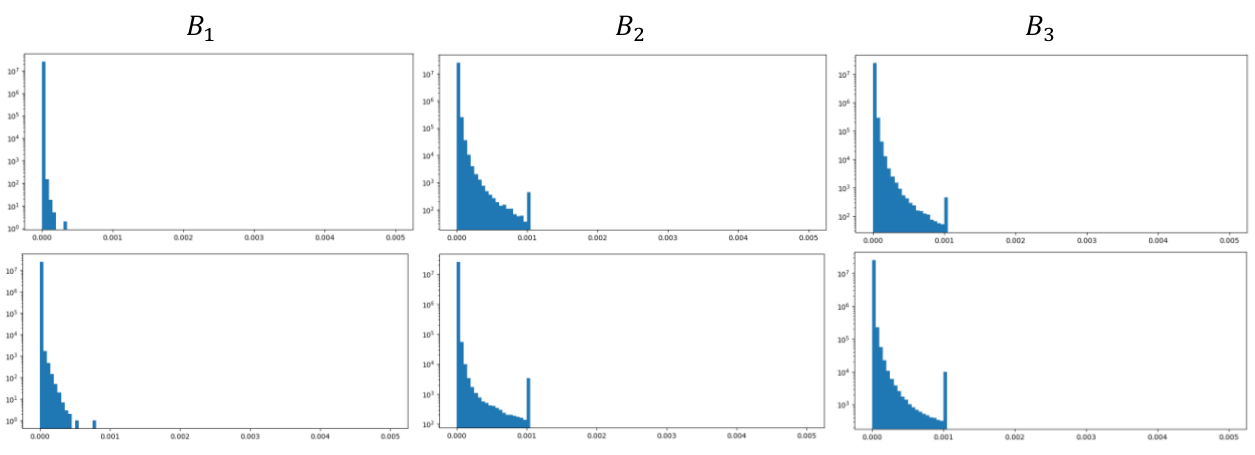}
   \caption{CaffeNet trained on CORe50 SIT. The first row shows $\hat{F}$ values distribution obtained by SI on batches $B_1, B_2$ and $B_3$. $\hat{F}$ values distribution from EWC is reported in the second row for comparison. The shape of the distribution is quite similar, even if in this experiments, the number of SI saturated values is about 10 times lower.}
   \label{fig:fvalues_si}
\end{figure}

Given $\hat{F}$ values, SI regularization can be implemented as EWC. The magnitude of $\hat{F}$ values can also be made comparable to EWC by proper setting of $w_i$, so that $max_F$ and $\lambda$ can take similar values. Figure \ref{fig:fvalues_si} compares the distributions of $\hat{F}_k$ values between EWC and SI: at first glance the distributions appear to be similar; of course more precise correlation studies could be performed, but this is out of the scope of this work.   

Summarizing, SI implementation is quite simple and, for each batch $B_i$, its overhead consists of \texttt{i)} computation of weight importance $F^i$, based on information already available during SGD; \texttt{ii)} storage of $F$ and $\Theta$, totaling $2 \cdot m$ values, where $m$ is the number of model weights.

\section{AR1: Combining Architectural and Regularization Strategies}
\label{sec:ar1_combining_architectural_and_regularization_strategies}

In this section we introduce AR1, a novel continuous learning approach obtained by combining architectural and regularization strategies. We start by reviewing CWR, a simple architectural technique that we introduced in \cite{pmlr-v78-lomonaco17a} and that, in spite of its simplicity, proved to be effective in SIT scenario. Then we first improve CWR with two simple but valuable modifications (that is, mean-shift and zero initialization) and finally extend it by combining it with SI. 

\subsection{Copy Weight with Reinit (CWR)}

CWR was proposed in \cite{pmlr-v78-lomonaco17a} as baseline technique for continuous learning from sequential batches. While it can work both for NC (new classes) and NIC (new instances and classes) update content type, here we focus on NC under SIT scenario.

Referring to Figure \ref{fig:mt_vs_sit} (bottom) the most obvious approach to implement a SIT strategy seems to be:
\begin{enumerate}
	\item Freeze shared weights $\bar{\Theta}$ after the first batch.
	\item For each batch $B_i$, extend the output layers with new neurons/weights for the new classes, randomly initialize the new weights but retain the optimal values for the old class weights. The old weights could then be frozen (denoted as FW in \cite{Lomonaco2017-arxiv}) or continued to be tuned (denoted as CW in \cite{Lomonaco2017-arxiv}).
\end{enumerate}

Implementing step 2 as above proved to be suboptimal with respect to CWR approach (see Figure 8 of \cite{Lomonaco2017-arxiv} for a comparison) where old class weights are re-initialized at each batch. 

\begin{algorithm}[!htb]
\caption{CWR}\label{algo:cwr}
\begin{algorithmic}[1]

\State $cw=0$
\State init $\bar{\Theta}$ random or from a pre-trained model (e.g. ImageNet)
\State for each training batch $B_i$:
\State \ \ \ \ \  expand output layer with $s_i$ neurons for the new classes in $B_i$
\State \ \ \ \ \ random re-init $tw$ (for all neurons in the output layer) 
\State \ \ \ \ \ Train the model with SGD on the $s_i$ classes of $B_i$:
\State \ \ \ \ \ \ \ \ \ \ if $B_i = B_1$ learn both $\bar{\Theta}$ and $tw$ 
\State \ \ \ \ \ \ \ \ \ \ else learn $tw$ while keeping $\bar{\Theta}$ fixed
\State \ \ \ \ \ for each class $j$ among the $s_i$ classes in $B_i$:
\State \ \ \ \ \ \ \ \ \ \ $cw[j] = w_i \cdot tw[j]$
\State \ \ \ \ \ Test the model by using $\bar{\Theta}$ and $cw$ 

\end{algorithmic}
\end{algorithm}

To learn class-specific weights without interference among batches, CWR maintains two sets of weights for the output classification layer: $cw$ are the consolidated weights used for inference and $tw$ the temporary weights used for training: $cw$ are initialized to 0 before the first batch, while $tw$ are randomly re-initialized (e.g., Gaussian initialization with std = 0.01, mean = 0) before each training batch. At the end of each batch training, the weights in $tw$ corresponding to the classes in the current batch are scaled and copied in $cw$: this is trivial in NC case because of the class segregation in different batches but is also possible for more complex cases (see Section 5.3 of \cite{pmlr-v78-lomonaco17a}). To avoid forgetting in the lower levels, after the first batch $B_1$, all the lower level weights $\bar{\Theta}$ are frozen. Weight scaling (with batch specific weights $w_i$) is necessary in case of unbalanced batches with respect to the number of classes or number of examples per class.

More formally, let $cw[j]$ and $tw[j]$ be the subset\footnote{the number of weights in each subset typically corresponds to the number of neurons in the penultimate layer.} of weights related to class $j$, then CWR learning sequence can be implemented as described in Algorithm \ref{algo:cwr}.

In CWR experiments reported in \cite{pmlr-v78-lomonaco17a}, to better disentangle class-specific weights we used models without class-shared fully connected layers (e.g., we removed FC6 and FC7 in CaffeNet). In fact, since $\bar{\Theta}$ weights are frozen after the first batch, fully connected layer weights tend to specialize on the first batch classes only. However, since skipping fully connected layers is not mandatory for CWR, in this paper to better compare different approaches we prefer to focus on native models and keep fully connected layers whether they exist.
Finally, CWR implementation is very simple and, the extra computation is negligible and for each batch $B_i$, its overhead consists of the storage of temporary weights $tw$, totaling $s \cdot pn$ values, where $s$ is the number of classes and $pn$ the number of penultimate layer neurons.

\subsection{Mean-shift and Zero Initialization (CWR+)}
\label{sub:mean_shift_and_zero_initialization_cwr_}

Here we propose two simple modifications of CWR: the resulting approach is denoted as CWR+. 
The first modification, mean-shift, is an automatic compensation of batch weights $w_i$. In fact, tuning such parameters is annoying and a wrong parametrization can lead the model to underperform. We empirically found that, if the weights $tw$ learnt during batch $B_i$, are normalized by subtracting their global average, then rescaling by $w_i$ is no longer necessary (i.e., all $w_i = 1$). Other reasonable forms or normalization, such as setting standard deviation to 1, led to worse results in our experiments. 

\begin{algorithm}[!htb]
\caption{CWR+}\label{algo:cwr+}
\begin{algorithmic}[1]

\State $cw=0$
\State init $\bar{\Theta}$ random or from a pre-trained model (e.g. ImageNet)
\State for each training batch $B_i$:
\State \ \ \ \ \ expand output layer with $s_i$ neurons for the new classes in $B_i$
\State \ \ \ \ \ $\mathbf{tw = 0}$ (for all neurons in the output layer)
\State \ \ \ \ \ Train the model with SGD on the $s_i$ classes of $B_i$:
\State \ \ \ \ \ \ \ \ \ \ if $B_i = B_1$ learn both $\bar{\Theta}$ and $tw$ 
\State \ \ \ \ \ \ \ \ \ \ else learn $tw$ while keeping $\bar{\Theta}$ fixed
\State \ \ \ \ \ for each class $j$ among the $s_i$ classes in $B_i$:
\State \ \ \ \ \ \ \ \ \ \ $\mathbf{cw[j] = tw[j] - avg(tw)}$
\State \ \ \ \ \ Test the model by using $\bar{\Theta}$ and $cw$ 

\end{algorithmic}
\end{algorithm}

The second modification, denoted as zero init, consists in setting initial weights $tw$ to 0 instead of typical Gaussian or Xavier random initialization. It is well known that neural network weights cannot be initialized to 0, because this would cause intermediate neuron activations to be 0, thus nullifying back-propagation effects. While this is certainly true for intermediate level weights, it is not the case for the output level (see Appendix \ref{sec:on_initializing_output_weight_to_zero} for a simple derivation). Actually, what is important here is not using the value 0, but the same value for all the weights: 0 is used for simplicity. Even if this could appear a minor detail, we discovered that it has a significant impact on the training dynamic and the forgetting. If output level weights are initialized with Gaussian or Xavier random initialization they typically take small values around zero, but even with small values in the first training iterations the softmax normalization could produce strong predictions for wrong classes. This would trigger unnecessary errors backpropagation changing weights more than necessary. While this initial adjustment is uninfluential for normal batch training we empirically found that is detrimental for continuous learning. Even if in this paper we apply zero init only to CWR+ and AR1, we found that even a simple approach such as fine tuning can greatly benefit from zero init in case of continuous learning.

In Algorithm \ref{algo:cwr+} we report the pseudocode for CWR+: the modifications w.r.t. CWR are highlighted in bold. CWR+ overhead is basically the same of CWR since taking the average of $tw$ is computationally negligible w.r.t. the SGD complexity.

\subsection{AR1}

A drawback of CWR and CWR+ is that weights $\bar{\Theta}$ are tuned during the first batch and then frozen. AR1, is the combination of an Architectural and Regularization approach. In particular, we extend CWR+ by allowing $\bar{\Theta}$ to be tuned across batches subject to a regularization constraint (as per LWF, ECW or SI). We did several combination experiments on CORe50 to select a regularization approach; each approach required a new hyperparameter tuning w.r.t. the case when it was used in isolation. At the end, our choice for AR1 was in favor of SI because of the following reasons:

\begin{itemize}
	\item LWF performs nicely in isolation, but in our experiments it does not bring relevant contributions to CWR+. We guess that being the LWF regularization driven by an output stability criterion, most of the regularization effects go to the output level that CWR+ manages apart. 
	\item Both EWC and SI provide positive contributions to CWR+ and their difference is minor. While SI can be sometime unstable when operating in isolation (see Sections \ref{sub:results_on_core50} and \ref{sub:results_on_icifar_100}) we found it much more stable and easy to tune when combined with CWR+. 
	\item SI overhead is small, since the computation of trajectories can be easily implemented from data already computed by SGD.
\end{itemize}

In Algorithm \ref{algo:ar1} we report the pseudocode for AR1.

\begin{algorithm}[!htb]
\caption{AR1}\label{algo:ar1}
\begin{algorithmic}[1]

\State $cw=0$
\State init $\bar{\Theta}$ random or from a pre-trained model (e.g. ImageNet)
\State $\Theta=0$ ($\Theta$ are the optimal shared weights resulting from the last training, see Section \ref{sub:synaptic_intelligence})
\State $\hat{F}=0$ ($\hat{F}$ is the weight importance matrix, see Section \ref{sub:synaptic_intelligence}).
\State for each training batch $B_i$:
\State \ \ \ \ \ expand output layer with $s_i$ neurons for the new classes in $B_i$
\State \ \ \ \ \ $tw = 0$ (for all neurons in the output layer)
\State \ \ \ \ \ Train the model with SGD on the $s_i$ classes of $B_i$ by simultaneously:
\State \ \ \ \ \ \ \ \ \ \ learn $tw$ with no regularization
\State \ \ \ \ \ \ \ \ \ \ learn $\bar{\Theta}$ subject to SI regularization according to $\hat{F}$ and $\Theta$
\State \ \ \ \ \ for each class $j$ among the $s_i$ classes in $B_i$:
\State \ \ \ \ \ \ \ \ \ \ $cw[j] = tw[j] - avg(tw)$
\State \ \ \ \ \ $\Theta = \bar{\Theta}$
\State \ \ \ \ \ Update $\hat{F}$ according to trajectories computed on $B_i$ (see eq. \ref{eq:Fk} and \ref{eq:fhat})
\State \ \ \ \ \ Test the model by using $\bar{\theta}$ and $cw$ 

\end{algorithmic}
\end{algorithm}

AR1 overhead is the sum of CWR+ and SI overhead:
\begin{itemize}
	\item storage:
		\begin{itemize}
			\item Temporary weights $tw$, totaling $s \cdot pn$ values, where $s$ is the number of classes and $pn$ the number of penultimate layer neurons.
			\item $F$ and $\Theta$, totaling $2 \cdot (m-s\cdot pn)$, where $m$ is the total number of model weights.
		\end{itemize}
	\item computation:
			\begin{itemize}
			\item Weights importance $\hat{F}$, based on information already available during SGD.
			\item Learning $sw$ subject to SI regularization can be easily implemented as weight decay (see eq. \ref{eq:wheights_update}) and is computationally light.
		\end{itemize}
\end{itemize}

Considering the low computational overhead and the fact that typically SGD is typically early stopped after 2 epochs, AR1 is suitable for online implementations. 

\section{Experiments}
\label{sec:experiments}

\subsection{CNN Models for CORe50}
\label{sub:cnn_models_for_core50}

\begin{figure}[!htb]
  \centering
  \includegraphics[width=0.8\textwidth]{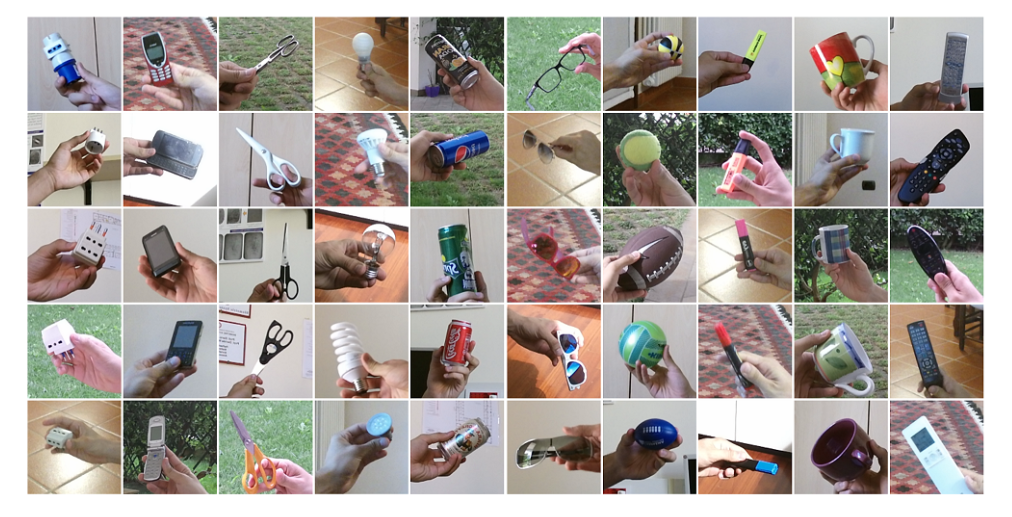}
  \caption{Example images of the 50 objects in CORe50. Each column denotes one of the 10 categories, but for the experiments reported in this paper we associate each object to a distinct class.}
  \label{fig:core50}
\end{figure}

The experiments reported in this paper have been carried out with two CNN architectures: CaffeNet \cite{Jia2014} and GoogLeNet \cite{Szegedy2014}. Minor modifications have been done to the default models as here detailed:

\begin{itemize}
	\item CaffeNet and GoogLeNet original models work on input images of size $227\times227$ and $224\times224$ respectively. CORe50 images are $128\times128$ and stretching them to $227\times227$ or $224\times224$ is something that should be avoided (would increase storage and amount of computation). On the other hand, as discussed in appendix A of \cite{Lomonaco2017-arxiv}, simply reshaping the network input size leads to a relevant accuracy drop. This is mainly due to the reduced resolution of feature maps whose size along the hierarchy is about half the original. We noted that the accuracy can be restored by halving the stride in the first convolutional layer and by adjusting padding if necessary: unfortunately, this also restore much of the original computational complexity, but \texttt{i)} does not require unnecessary image stretching, \texttt{ii)} allows to save memory, and \texttt{iii)} reduces CPU$\rightarrow$GPU data transfer when loading mini-batches. 
	\item For CaffeNet, the number of neurons in fc6 and fc7 fully connected layers was halved. This substantially reduce the number of parameters without any relevant accuracy drop.
	\item GoogLeNet has three output layers. The deepest one is typically used for prediction, while the two intermediate ones are useful to boost gradient flow during training thank to the injection of a fresh supervised signals at intermediate depth. While the deepest output level is proceeded by a global average pooling, each intermediate output layer is preceded by a fully connected layer; in our experiment where GoogLeNet was always initialized from ImageNet such fully connected layers did not provide any advantage, hence to simplify the architecture and reduce the number of parameters we removed them. Finally, note that when GoogLeNet is used with CWR, CWR+, and AR1 we need to maintain in $cw$ a copy of the weights of all the three output levels.
\end{itemize}

Table \ref{tab:cnn_variations} in Appendix \ref{sec:architectural_changes_in_the_models_used_on_core50} lists all the changes between original and modified models. Model weights (in convolutional layers) have been always initialized from ImageNet pre-training.
We also tried other popular CNN architectures on CORe50, including NiN \cite{Lin2014} and ResNet-50 \cite{He2015}. Table \ref{tab:cnn_models} compares the accuracy of these models when trained on the whole training set (i.e., all training batches combined). The gap between GoogLeNet and ResNet50 is quite small, but the latter is much slower to train, so we decided to use GoogLeNet  as a near state-of-the-art model.

\begin{table}[!htb]
  \caption{Model training on CORe50 by using the whole training set (fusion of all training batches). Models are adapted to work with 128x128 inputs and weights in convolutional layers are initialized from ImageNet pre-training. Time refers to a single Titan X Pascal GPU and Caffe framework.}
  \label{tab:cnn_models}
  \centering
  \begin{tabular}{lllll}
    \toprule
    \textbf{Model (128x128)}     & \textbf{Test accuracy}     & \textbf{Mini-batch size}   & \textbf{\# Epochs} & \textbf{Time (m)} \\
    \midrule
    CaffeNet 			& 74.1\%  			& 256				& 4			& 7     \\
    NiN     			& 82.3\% 			& 256 				& 4			& 14      \\
    GoogLeNet  			& 91.3\%       		& 64				& 4			& 30  \\
    ResNet-50 			& 92.9\%  			& 12				& 4			& 120     \\
    \bottomrule
  \end{tabular}
\end{table}

\subsection{Results on CORe50}
\label{sub:results_on_core50}

CORe50 dataset \cite{pmlr-v78-lomonaco17a} was specifically designed as a benchmark for continuous object recognition (Figure \ref{fig:core50}). Here we consider NC content update type (a.k.a. incremental-class learning) where the 50 classes are partitioned in 9 batches provided sequentially: $B_1$ includes 10 classes while $B_2, \dots, B_9$ 5 classes each. For each class 2400 patterns (300 frames $\times$ 9 sessions) are included in the training batches and 900 patterns (300 frames $\times$ 3 sessions) are segregated in the test set. The test set is fixed and the accuracy is evaluated after each batch on the whole test set, including still unseen classes. Refer to Section 5.2 of \cite{pmlr-v78-lomonaco17a} for a discussion about CORe50 testing protocol. Experiments in Section \ref{sub:results_on_icifar_100} confirm the usefulness of reporting results on a fixed test set.

\begin{figure}[!htb]
  \centering
  \includegraphics[width=\textwidth]{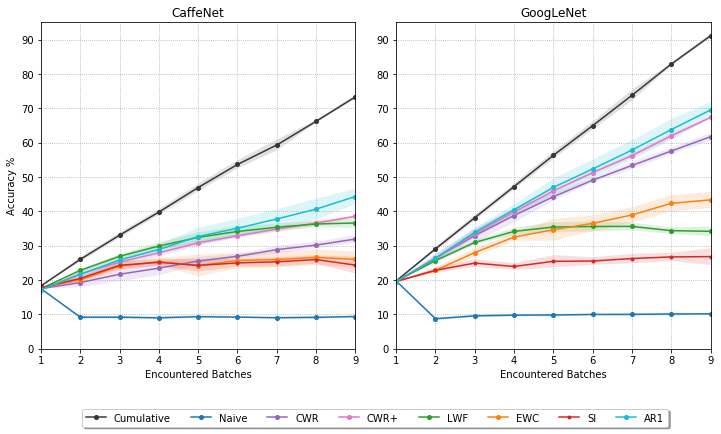}
  \caption{The graphs show CaffeNet and GoogLeNet accuracy on CORe50 after each training batch (average on 10 runs, with different class ordering). Colored areas represent the standard deviation of each curve. Better viewed in color.}
  \label{fig:core50_sit_results}
\end{figure}

Continuous learning approaches introduced in Section \ref{sec:continuous_learning_strategies_in_sit} and \ref{sec:ar1_combining_architectural_and_regularization_strategies} are here compared with two baseline approaches:
\begin{enumerate}
	\item \textbf{Naive}: simply tunes the model across the batches without any specific mechanism to control forgetting, except early stopping.
	\item \textbf{Cumulative}: the model is retrained with the patterns from the current batch and all the previous batches (only for this approach we assume that all previous data can be stored and reused). This is a sort of upper bound, or ideal performance that continuous learning approaches should try to reach.
\end{enumerate}

LWF, EWC and SI were run with the modifications (variable lambda, normalization, clipping, etc.) introduced in Section \ref{sec:continuous_learning_strategies_in_sit}. In fact, when the approaches were tested in their native form, either we were not able to make them consistently learn across the batches or to contrast forgetting enough.   

Figure \ref{fig:core50_sit_results} shows CaffeNet and GoogLeNet accuracy in CORe50, SIT - NC scenario. Accuracy is the average over 10 runs with different batch ordering. For all the approaches hyperparameter tuning was performed on run 1, and then hyperparameters were fixed across runs $2, \dots, 10$. Table \ref{tab:hyper_core50} in Appendix \ref{sec:hyperparameter_values_for_core50} shows hyperparameter values for all the methods. From these results we can observe that: 

\begin{itemize}
	\item The effect of catastrophic forgetting is well evident for the \textbf{Naïve} approach: accuracy start from 17-20\% at the end of $B_1$ (in line with the 20\% of classes in $B_1$), but then suddenly drops to about 9-10\% (that is, the proportion of classes in each subsequent batch); in other words, as expected, after each batch, the model tends to completely forget previous classes while learning the new ones.  
	\item \textbf{LWF} behaves well for CaffeNet and moderately well for GoogLeNet: continuous learning is evident and accuracy is much better than Naive approach. A few percentage point drop can be noted in the last batches for GoogLeNet; to avoid this we tried to boost stability by increasing the lambda value: this yielded to an increasing learning trend across all the batches but the top accuracy was never higher than 32\% (neither in the central batches nor in the last ones), so we decided to rollback to previous parametrization. 
	\item Both the models are able to learn incrementally with \textbf{EWC}, but while for GoogLeNet the learning trend is quite good, CaffeNet after a few batches tends to stabilize and the final accuracy is much lower. Since GoogLeNet is a deeper and more powerful network we can expect a certain accuracy gap (see the corresponding cumulative approach curves); however here the gap is much higher than for LWF and in our experiment we noted that EWC (and SI) tend to suffer more than LWF the presence of fully connected layers such as FC6 and FC7 layer in CaffeNet\footnote{GoogleNet, as many modern network architectures, does not include fully connected layers before the last classification layer.}. Fully connected layers are usually close to the final classification layer, and any change in their weights is likely to have a major impact on classification.
	Even if EWC can operate on all layers, precisely constraining the weights of fully connected layers appears to be challenging due to their high importance for all tasks. LWF, whose regularization signal ``comes from the top'', seems to better deal with fully connected layers. To further investigate this issue some experiments have been performed by removing FC6 and FC7 from CaffeNet and, in spite of the shallower and less powerful network, for EWC we got a few percentage improvements while LWF accuracy slightly decreased. 
	\item While CaffeNet accuracy with \textbf{SI} is very close to EWC, GoogLeNet accuracy with SI is markedly lower than EWC. In general, we noted that SI is less stable than EWC and we believe that SI weights importance estimation can be sometimes less accurate than the EWC one because of the following reasons:
	\begin{itemize}
		\item A weight which is not moved by SGD is not considered important for a task by SI, but this is not always true. Let us assume that Batch $B_1$ trains the model on classes $c_1, c_2, \dots, c_{10}$; the output layer weights which are not connected to the above class output neurons, are probably left unchanged to their (near 0) initial value; however, this does not mean that they are not important for $B_1$ because if we raise their values the classification might dramatically change. More in general, if a weight already has the right value for a task and is not moved by SGD, concluding that it is not important for the task can be sometime incorrect.
		\item A weight could cover a long trajectory and, at the end of the training on a batch, assume a value similar to the initial one (closed trajectory). Such situation can happen because the loss surface is highly non convex and gradient descent could increase a weight while entering a local minimum and successively restoring its value once escaped.  
	\end{itemize}
	\item \textbf{CWR} and \textbf{CWR+} accuracy is quite good, always better than LWF, EWC, SI on GoogLeNet, and better than EWC and SI on CaffeNet. Continuous learning trend is here nearly linear across batches, with no evident saturation effect. The relevant improvement of CWR+ over CWR is mainly due to zero init.
	\item \textbf{AR1} exhibits the best performance. SI regularization is here quite stable and pushes up CWR+ accuracy. AR1 is also stable w.r.t. parameter tuning: in Table \ref{tab:hyper_core50} one can observe that we used almost the same hyperparameters for CaffeNet and GoogLeNet. For GoogleNet AR1 reaches a remarkable accuracy of about 70\% with small standard deviation among runs, and the gap w.r.t. the Cumulative approach is not large, thus proving that continuous learning (without storing/reusing old patterns) is feasible in a complex incremental class scenario.
\end{itemize}

\subsection{Results on iCIFAR-100}
\label{sub:results_on_icifar_100}

CIFAR-100 \cite{Krizhevsky2009} is a well-known and largely used dataset for small ($32\times32$) natural image classification. It includes 100 classes containing 600 images each (500 training + 100 test). The default classification benchmark can be translated into a SIT-NC scenario (denoted as iCIFAR-100 by \cite{Rebuffi2017}) by splitting the 100 classes in groups. In this paper we consider groups of 10 classes thus obtaining 10 incremental batches. 

\begin{figure}[!htb]
  \centering
  \includegraphics[width=\textwidth]{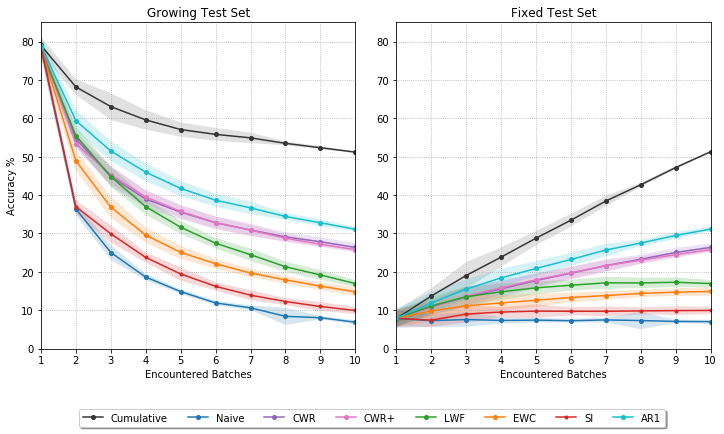}
  \caption{Accuracy on iCIFAR-100 with 10 batches (10 classes per batch). Results are averaged on 10 runs: for all the strategies hyperparameters have been tuned on run 1 and kept fixed in the other runs. The experiment on the right, consistently with CORe50 test protocol, considers a fixed test set including all the 100 classes, while on the left we include in the test set only the classes encountered so far (analogously to results reported in \cite{Rebuffi2017}). Colored areas represent the standard deviation of each curve. Better viewed in color.}
  \label{fig:icifar_sit_results}
\end{figure}

The CNN model used for this experiment is the same used by \cite{Zenke2017} for experiments on CIFAR-10/100 Split and whose results have been reported in Figure \ref{fig:cifar_split_res} for the MT scenario. It consist of 4 convolutional + 2 fully connected layers; details are available in Appendix A of \cite{Zenke2017}. The model was pre-trained on CIFAR-10 \cite{Krizhevsky2009}. Figure \ref{fig:icifar_sit_results} compares the accuracy of the different approaches on iCIFAR-100. The results obtained on CORe50 are almost confirmed, in particular:

\begin{itemize}
	\item Unlike the \textbf{Naïve} approach, \textbf{LWF} and \textbf{EWC} provide some robustness against forgetting, even if in this incremental scenario their performance is not satisfactory. \textbf{SI}, when used in isolation, is quite unstable and performs worse than LWF and EWC.
	\item The accuracy improvement of \textbf{CWR+} over \textbf{CWR} is here very small, because the batches are balanced (so weight normalization is not required) and the CNN initialization for the last level weights was already very close to 0 (we used the authors’ default setting of a Gaussian with std = 0.005).
	\item \textbf{AR1} consistently outperforms all the other approaches.
\end{itemize}

It is worth noting that both the experiments reported in Figure \ref{fig:icifar_sit_results} (i.e., fixed and expanding test set) lead to the same conclusions in terms of relative ranking among approaches, however we believe that a fixed test set allows to better appreciate the incremental learning trend and its peculiarities (saturation, forgetting, etc.) because the classification complexity (which is proportional to the number of classes) remains constant across the batches. For example, in the right graph it can be noted that SI, EWC and LWF learning capacities tend to saturate after 6-7 batches while CWR, CWR+ and AR1 continue to grow; the same information is not evident on the left because of the underlying negative trend due to the increasing problem complexity.

Finally note that absolute accuracy on iCIFAR-100 cannot be directly compared with \cite{Rebuffi2017} because the CNN model used in \cite{Rebuffi2017} is a ResNet-32, which is much more accurate than the model here used: on the full training set the model here used achieves about 51\% accuracy while ResNet-32 about 68,1\% \cite{Pan2018}.

\subsection{Preliminary Comparison with Rehearsal-based Approaches}
\label{subsec:rehe_comp}

While the focus of the paper, as pointed out in Section \ref{sec:continuous_learning_strategies_in_sit}, is on architectural/regularization strategies (i.e., strategies not storing any past input data), in this section we present a preliminary comparison of AR1 with three representative rehearsal-based approaches: basic rehearsal\footnote{The external memory is filled by sampling past examples from both the external memory and the current training batch. The sampling probability is tuned to balance the number of samples drawn from different batches.} (REHE), GEM \cite{Lopez-paz2017} and ICARL \cite{Rebuffi2017}. Results contained in this section represent a preliminary assessment, since we did not re-implement GEM and ICARL as the other regularization techniques but adapted the implementations made available by their authors to the CORe50 benchmark and the SIT-NC scenario leaving most of the hyperparameters and implementation choices unchanged. In particular, GEM was specifically designed for a Multi-Task setting, and, to the best of our knowledge these are the first experiments on more complex benchmarks such as CORe50 and a Single-Incremental-Task scenario with images larger than 32x32.

\begin{figure}[!htb]
  \centering
    \subfloat{{\includegraphics[width=6.9cm]{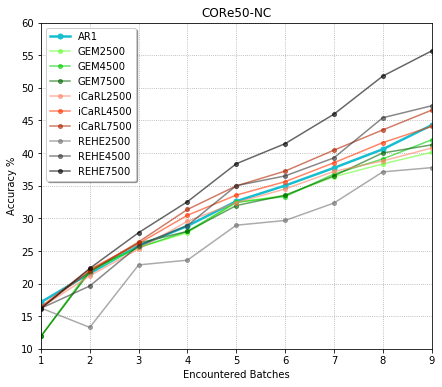} }}%
    \subfloat{{\includegraphics[width=6.9cm]{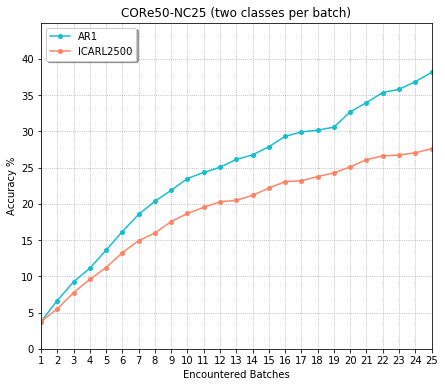} }}%
  \caption{On the left, a comparison of AR1 with rehearsal strategies with different amount of external memory (2500, 4500 and
7500 input patterns). Accuracy is averaged over 3 runs. On the right, a comparison of AR1 with ICARL on a more complex setting
where the original 50 classes are split in 25 batches with 2 classes each. For this experiment we also computed the backward
transfer metric \cite{Lopez-paz2017} which is -15,64\% for AR1 and -11,09\% for ICARL.}
  \label{fig:rehe_comp}
\end{figure}

In Figure \ref{fig:rehe_comp} (left), we report the average accuracy of 3 rehearsal-based strategies: REHE, GEM and ICARL each with external memory size of 2500, 4500 and 7500 input patterns (corresponding to about 2\%, 4\% and 6\% of the total training images). We note that AR1, while not storing any training example from the past, is competitive with respect the other approaches, especially when a high memory and computational budget is not available. It is worth noting that the external memory is used differently by ICARL and GEM; in fact, while ICARL is designed to fill the total storage after every batch, GEM uses a fixed amount of memory for
each batch and reaches the total memory budget only at the end of the last batch.

In Figure \ref{fig:rehe_comp} (right), we compare AR1 an ICARL on a more difficult setting with respect to the original CORe50-NC benchmark; the 50 classes are here split in 25 batches with 2 classes each. It is interesting to observe that even if the more challenging problem leads to a certain degradation in terms of absolute accuracy, both techniques perform reasonably well also with smaller batches.

\FloatBarrier
\section{Practical Advices for Hyperparameters Tuning}
\label{sec:practical_advices_for_hyperparameters_tuning}

Hyperameters tuning is not easy for complex deep architectures and is still more complex for continuous learning over sequential batches. Simply looking at the accuracy trend along the batches is not enough to understand if an approach is properly parametrized. For example, a poor accuracy can be due to insufficient learning of the new classes or by the forgetting of the old ones. 

\begin{figure}[!htb]
  \centering
  \includegraphics[width=\textwidth]{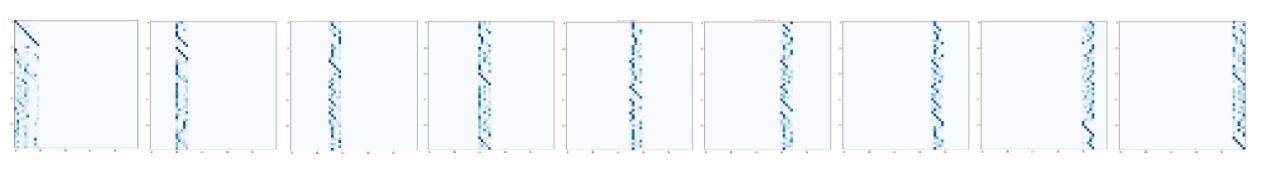}
  \caption{Sequence of confusion matrices computed after each training batch for the Naïve approach on CaffeNet. On the vertical axis the true class and on the abscissa the predicted class.}
  \label{fig:cms_naive}
\end{figure}

\begin{figure}[!htb]
  \centering
  \includegraphics[width=\textwidth]{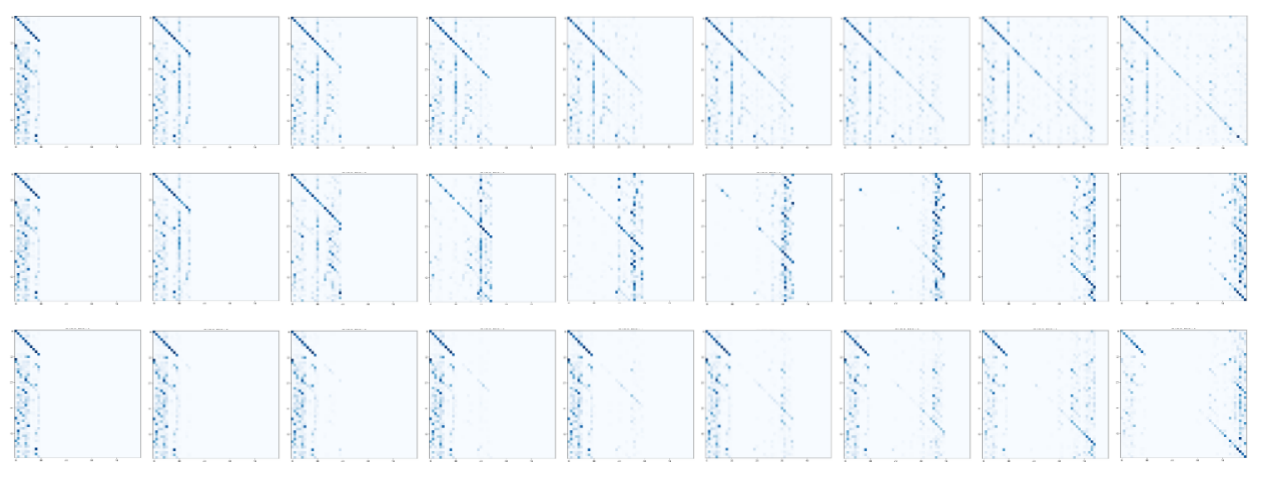}
  \caption{Sequence of confusion matrices computed after each training batch ($1, \dots, 9$) for the LWF approach and CaffeNet. In the first row the approach is properly parametrized (variable $\lambda$ and $map$ function, see Table \ref{tab:hyper_core50}) and the model is able to continuously learn new classes without forgetting the old ones. In the second we used the same $\lambda_i=0.5$ for all the batches: for $B_2, \dots, B_3$ the regularization is appropriate but for successive batches is too light leading to excessive forgetting. In the third row we used the same $\lambda_i=0.8$ for all the batches: for $B_2, \dots, B_6$ the regularization is too strong and learning of corresponding classes is poor.}
  \label{fig:cms_lwf}
\end{figure}

\begin{figure}[!htb]
  \centering
  \includegraphics[width=0.6\textwidth]{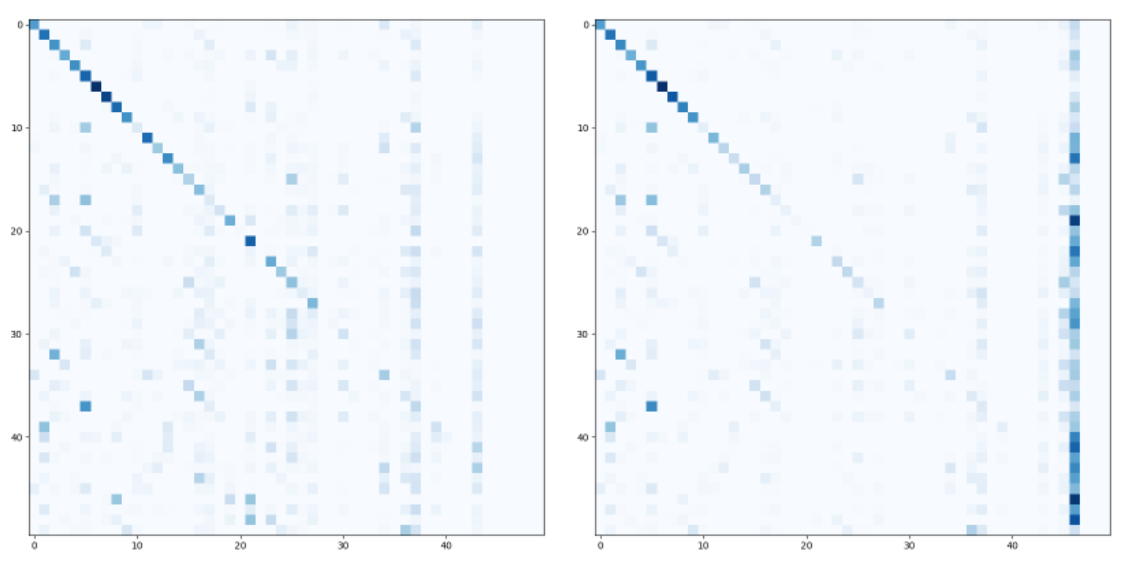}
  \caption{Confusion matrices after computed after $B_8$ and $B_9$ for the EWC approach and CaffeNet. As discussed in Section \ref{sub:results_on_core50} EWC tends to saturate CaffeNet capacity after the first 5-6 batches. In this specific run the amount of training on the last batch is too high w.r.t. the residual model capacity and the learning result are poor: the sharp vertical band on the right is an alarm signal.}
  \label{fig:cms_ewc}
\end{figure}

In our experience visualizing the confusion matrices (CM) is very important to understand what is happening behind the scenes. Looking at the last CM (that is after the last batch) is often not sufficient and the entire CM sequence must be considered. Figure \ref{fig:cms_naive} shows the CM sequence (one after each batch) for the naïve approach: forgetting is clearly highlighted by a vertical band moving from the left to the right to cover the classes of the most recent batch. Figure \ref{fig:cms_lwf} shows three CM sequences for LWF approach on CaffeNet: \texttt{i)} in the first row parametrization is good; \texttt{ii)} in the second row the model forget to much old classes, regularization should be increases; \texttt{iii)} in the third row initial regularization is too strong and the model cannot learn classes in the corresponding batches. 

\begin{figure}[!htb]
  \centering
  \includegraphics[width=\textwidth]{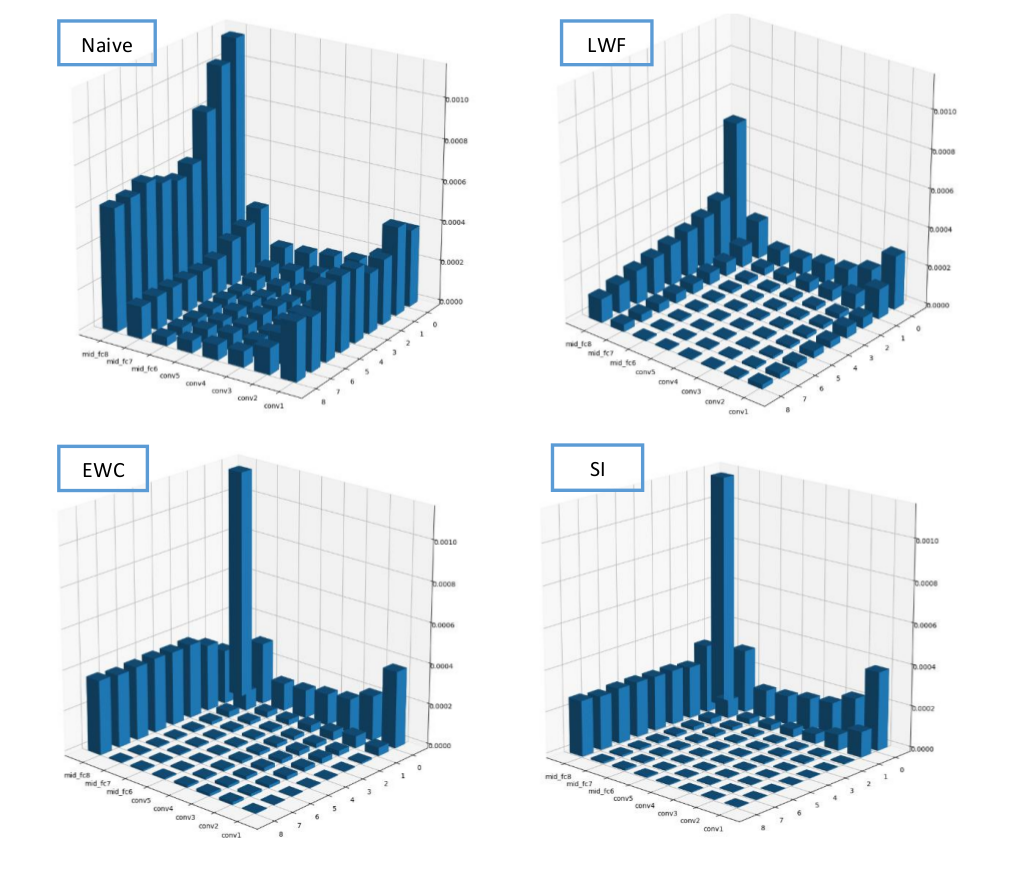}
  \includegraphics[width=\textwidth]{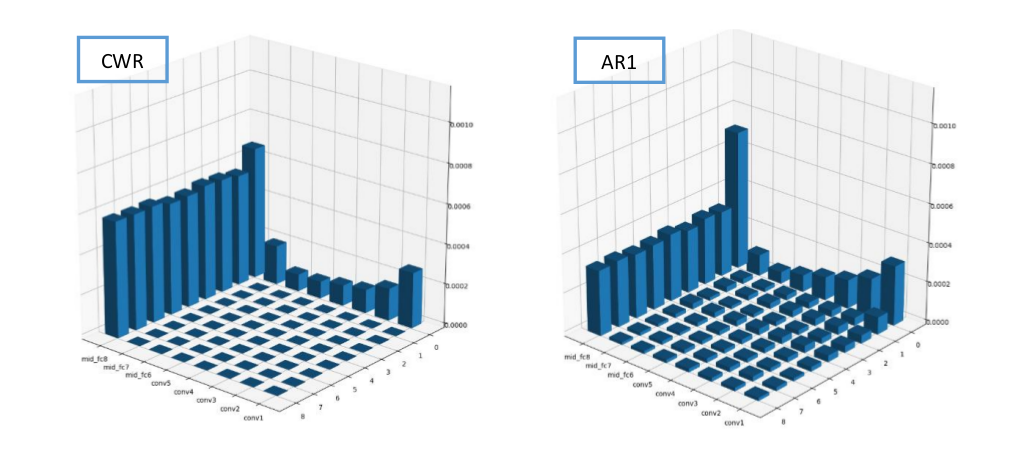}
  \caption{Amount of weight changes by layer and training batch in CaffeNet for different approaches.}
  \label{fig:3d_changes}
\end{figure}

Unfortunately, the trade-off stability/plasticity does not depends only on the regularization strength (e.g. $\lambda_i$ for LWF) because the learning rate and the number of training epochs are also indirectly related. However, looking at the CM sequence allows to understand what is the direction of change of one or more parameters. It can also happen that the amount of regularization is good for the first batches, but unsatisfactory for the successive ones; the CM sequence easily reveals this and allows to take countermeasures (e.g. the $map$ function for LWF). Finally, when a model is strongly regularized its learning capacity tends to saturate (e.g. in EWC there are no ``free'' parameters to move in order to learn new classes); this is usually reflected by anomalies such as sharp vertical bars in correspondence of one or few classes (see Figure \ref{fig:cms_ewc}).

Another useful diagnostic technique is visualizing (in the form of a 3D histogram) the average amount of change of the weights in each layer at the end of each batch. To avoid cancellation due to different sign, we compute the average of absolute values. Figure \ref{fig:3d_changes} shows the histograms for CaffeNet. We can observe that in the naïve approach weights (which are not constrained) are significantly changed throughout all the layers for all batches. On the other hand, in the regularization approaches, changes tend to progressively reduce batch by batch, and most of the chances occur in the top layers. While in CWR the $\bar{\Theta}$ freeze is well evident, in AR1 intermediate layers weights are moved (more than in EWC and SI) without negative impact in terms of forgetting.

\section{Conclusions}
\label{sec:conclusions}

In this paper we introduced a novel approach (AR1) for SIT scenario combining architectural and regularization strategies. Results on CORe50 and iCIFAR-100 proves that AR1 allows to train sequentially complex models such as CaffeNet and GoogLeNet by limiting the detrimental effects of catastrophic forgetting. 

AR1 accuracy was higher than existing regularization approaches such as LWF, EWC and SI and, from preliminary experiments, compares favorably with rehearsal techniques when the external memory size is not large. AR1 overhead in terms of storage is very limited and most of the extra computation is based on information made available by stochastic gradient descent. We showed that early stopping SGD after very few epochs (e.g., 2) is sufficient to incrementally learn new data on CORe50. Further ideas could be investigated in the future to quantify weight importance for old tasks such as exploiting the moving average of squared gradients already considered by methods like RMSprop \cite{Hinton2012a} or Adam \cite{Kingma2014} or the Hebbian-like reinforcements between active neurons recently proposed in \cite{Aljundi2017}.

In AR1, changing the underlying representation while the model learns new classes introduces some forgetting for the old ones; our experiments confirm that limiting the amount of change of important weights for the old classes, is quite effective to control forgetting. However, if we continue to add classes the network capacity tends to saturate and the representation cannot be further adapted to new data. To deal with this problem we believe that the incremental lateral expansion of some intermediate layers, similar to that proposed in \cite{Wanga} can be a valid approach. Class-incremental learning (NC update content type) is only one of the cases of interest in SIT. New instances and classes (NIC) update content type, available under CORe50, is a more realistic scenario for real applications, and would constitute the main target of our future research.

AR1 extension to unsupervised (or semi-supervised) implementations, such as those described in \cite{Maltoni2016-icpr} and \cite{Parisi2018a} is another scenario of interest for future studies. In particular, the 2-level self-organizing model (SOM) proposed by \cite{Parisi2018a} implements memory replay in a very effective way and is capable of exploiting temporal coherence in CORe50 videos with weak supervision. We believe that such a memory replay mechanism could be very effective in conjunction with AR1 and we are going to investigate if a similar mechanism could be embedded in a CNN without external SOM components.  

In conclusion, although much more validations in complex setting and new better approaches will be necessary, based on these preliminary results we can optimistically envision a new generation of systems and applications that once deployed continue to acquire new skills and knowledge, without the need of being retrained from scratch.

\FloatBarrier

\bibliography{library}
\bibliographystyle{plain}

\appendix

\section{Implementation Details (Caffe framework)}
\label{sec:implementation_details_caffe_framework_}

Since implementing dynamic output layer expansion was tricky in Caffe framework, we initially implemented the different strategies by using a single \emph{maximal head} (i.e., including all the problem classes since from the beginning) instead of an \emph{expanding head}. In principle, the two approaches are quite similar, since if a particular batch does not contain patterns of a given class, no relevant error signals are sent back along the corresponding connections during SGD. Hoverer, looking at the details of the training process, the two approaches are not exactly the same. 

For example, for CWR+ we verified, with some surprise, that the maximal head simplifying approach constantly leads to better accuracy (up to 6-7\% on CORe50) w.r.t. to the expanding head approach. We empirically found that the reason is related to the gradient dynamics during the initial learning iterations: working with a higher number of classes makes initial predictions smaller (because of softmax normalization) and the gradient correction for the true class stronger; in a second stage, predictions start to converge and the gradient magnitude is equivalent in the two approaches. It seems that for SGD learning (with fixed learning rate) boosting the gradient in the first iterations favors accuracy and reduces forgetting. We checked this by experimentally verifying that the expanding head approach combined with a variable learning rate performs similarly to maximal head with fixed learning rate. Therefore, to maximize accuracy and reduce complexity, CWR and its evolutions (CWR+ and AR1) have been implemented with the maximal output layer approach. Referring to the pseudocode in Algorithms \ref{algo:cwr}, \ref{algo:cwr+} and \ref{algo:ar1}, it is sufficient to keep  to constant maximum size (e.g., 50 for CORe50) and remove the line ``expand output layer with...''. 

For the other approaches we verified that: LWF performs slightly better with expanding head approach while EWC and SI work better (and are easy to tune) with maximal head. To produce the results presented in Section \ref{sec:experiments} we used for each strategy the approach that proved to be the most effective. Strategy specific notes are reported in the following for our Caffe implementation.

\paragraph{LWF.} It is worth noting that in Caffe a cross-entropy loss layer accepting soft target vectors is not available in the standard layer catalogue and a custom loss layer need to be created.

\paragraph{EWC.} To implement EWC in Caffe we:
\begin{itemize}
	\item compute, average and clip $F^i$ values in pyCaffe (for maximum flexibility). To calculate $F_k^i$ the variance of the gradient should be computed by taking the gradient of each of the $n_i$ patterns in isolation. To speed-up implementation and improve efficiency we computed the variance at mini-batch level, that is using the average gradients over mini-batches. In our experiment we did not note any performance drop even when using mini-batches of 256 patterns. 
	\item pass $F$ and $\Theta^*$ to the solver via a further input layer.
	\item modified SGD solver, by adding a custom regularization that performs EWC regularization in a weight decay style.
\end{itemize}
\paragraph{SI.} Starting from EWC implementation, SI can be easily setup in Caffe. In fact, the regularization stage is the same and we only need to compute $F^i$ values during SGD. For this purpose, in the current implementation, for maximum flexibility, we used pyCaffe. 

\section{Architectural Changes in the Models Used on CORe50}
\label{sec:architectural_changes_in_the_models_used_on_core50}

\begin{table}[!htb]
  \caption{Summary of changes w.r.t. the original CaffeNet and GoogLeNet models used in this paper.}
  \label{tab:cnn_variations}
  \centering
  \begin{tabular}{lll}
    \toprule
    \multicolumn{3}{c}{\textbf{CaffeNet}}                   \\
    \cmidrule(r){1-3}
    \multicolumn{1}{c}{\textit{Layer}}    & \multicolumn{1}{c}{\textit{Original}}    & \multicolumn{1}{c}{\textit{Modified}} \\
    \midrule
    data (Input) 			& size: $227\times227$  	& size: $128\times128$     \\
    conv1 (convolutional)   & stride: 4		 	& stride: 2      \\
    conv2 (convolutional)   & pad: 2     	 	& pad: 1  \\
    fc6 (fully connected)   & neurons: 4096     & neurons: 2048  \\
    fc7 (fully connected)   & neurons: 4096     & neurons: 2048  \\  
    fc8 (output)     		& neurons: 1000 (ImageNet classes)       & neurons: 50 (CORe50 classes)  \\  
    \toprule
    \multicolumn{3}{c}{\textbf{GoogLeNet}}   \\
    \cmidrule(r){1-3}
    \multicolumn{1}{c}{\textit{Layer}}    & \multicolumn{1}{c}{\textit{Original}}    & \multicolumn{1}{c}{\textit{Modified}} \\
    \midrule
    data (Input) 						& size: $224\times224$  					& size: $128\times128$     \\
    conv1/7x7\_s2 (convolutional)   	& stride: 2, pad: 3		 			& stride: 1, pad: 0      \\
    loss1/ave\_pool (pooling)   		& kernel: 5    	 					& kernel: 6  \\
    loss1/fc (fully connected)  		& neurons: 1024     				& layer removed  \\
    loss1/classifier (output int. 1)   	& neurons: 1000 (ImageNet classes)  & neurons: 50 (CORe50 classes)  \\  
    loss2/ave\_pool (pooling)     		& kernel: 5       					& kernel: 6  \\
    loss2/fc (fully connected)     		& neurons: 1024)       				& layer removed  \\
    loss2/classifier (output int. 2)    & neurons: 1000 (ImageNet classes)  & neurons: 50 (CORe50 classes)  \\
    loss3/classifier (output)     		& neurons: 1000 (ImageNet classes)  & neurons: 50 (CORe50 classes)  \\  
    \bottomrule

  \end{tabular}
\end{table}

\FloatBarrier
\section{Hyperparameter Values for CORe50}
\label{sec:hyperparameter_values_for_core50}

  \begin{longtable}{lll}
    \caption{the hyperparameter values used for CaffeNet and GoogLeNet on CORe50. The selection was performed on run 1, and hyperparameters were then fixed for runs $2, \dots,10$.}\\[-10pt]
  	\label{tab:hyper_core50}
  	\endfirsthead
	\endhead
    \toprule
    \multicolumn{3}{c}{\textbf{Cumulative}}                   \\
    \cmidrule(r){1-3}
    \multicolumn{1}{c}{\textit{Parameters}}    & \multicolumn{1}{c}{\textit{CaffeNet}}    & \multicolumn{1}{c}{\textit{GoogLeNet}} \\
    \midrule
    epochs, $\eta$ (learn. rate) 			& 4, 0.0025  	& 4, 0.0025     \\
    \bottomrule
    \pagebreak
    \toprule
    \multicolumn{3}{c}{\textbf{Naive}}   \\
    \cmidrule(r){1-3}
    \multicolumn{1}{c}{\textit{Parameters}}    & \multicolumn{1}{c}{\textit{CaffeNet}}    & \multicolumn{1}{c}{\textit{GoogLeNet}} \\
    \midrule
    Head (see App. A) 							& Maximal  				& Maximal     \\
    $B_1$: epochs, $\eta$ (learn. rate)   		& 2, 0.0003		 		& 4, 0.005      \\
    $B_i, i>1$: epochs, $\eta$ (learn. rate)   	& 2, 0.0003    	 		& 2, 0.0003  \\
    \toprule
    \multicolumn{3}{c}{\textbf{LWF}}   \\
    \cmidrule(r){1-3}
    \multicolumn{1}{c}{\textit{Parameters}}    & \multicolumn{1}{c}{\textit{CaffeNet}}    & \multicolumn{1}{c}{\textit{GoogLeNet}} \\
    \midrule
    Head (see App. A) 							& Expanding  			& Expanding     \\
    $map$										& [0.66...0.9] $\rightarrow$ [0.45...0.85] & [0.66...0.9] $\rightarrow$ [0.45...0.85] \\
    $B_1$: epochs, $\eta$ (learn. rate)   		& 2, 0.0003		 		& 4, 0.0003      \\
    $B_i, i>1$: epochs, $\eta$ (learn. rate)   	& 2, 0.0002    	 		& 2, 0.0002  \\
    \toprule
    \multicolumn{3}{c}{\textbf{EWC}}   \\
    \cmidrule(r){1-3}
    \multicolumn{1}{c}{\textit{Parameters}}    & \multicolumn{1}{c}{\textit{CaffeNet}}    & \multicolumn{1}{c}{\textit{GoogLeNet}} \\
    \midrule
    Head (see App. A) 							& Maximal  			& Maximal     \\
    $max_F$										& 0.001 			& 0.001  \\
    $\lambda$									& 5.0e7				& 3.4e7 \\
    $B_1$: epochs, $\eta$ (learn. rate)   		& 2, 0.001		 	& 4, 0.002      \\
    $B_i, i>1$: epochs, $\eta$ (learn. rate)   	& 2, 0.000025    	& 2, 0.000035  \\
    \toprule
    \multicolumn{3}{c}{\textbf{SI}}   \\
    \cmidrule(r){1-3}
    \multicolumn{1}{c}{\textit{Parameters}}    & \multicolumn{1}{c}{\textit{CaffeNet}}    & \multicolumn{1}{c}{\textit{GoogLeNet}} \\
    \midrule
    Head (see App. A) 							& Maximal  			& Maximal     \\
    $\xi$ 										& 1e-7  			& 1e-7     \\
    $w_1, w_i (i > 1)$ 							& 0.00001, 0.005  	& 0.00001, 0.005     \\
    $max_F$										& 0.001 			& 0.001  \\
    $\lambda$									& 5.0e7				& 3.4e7 \\
    $B_1$: epochs, $\eta$ (learn. rate)   		& 2, 0.001		 	& 4, 0.002      \\
    $B_i, i>1$: epochs, $\eta$ (learn. rate)   	& 2, 0.00002    	& 2, 0.000035  \\
    \toprule
    \multicolumn{3}{c}{\textbf{CWR}}   \\
    \cmidrule(r){1-3}
    \multicolumn{1}{c}{\textit{Parameters}}    & \multicolumn{1}{c}{\textit{CaffeNet}}    & \multicolumn{1}{c}{\textit{GoogLeNet}} \\
    \midrule
    Head (see App. A) 							& Maximal  			& Maximal     \\
    $w_1, w_i (i > 1)$ 							& 1.25, 1  			& 1, 1    \\
    $B_1$: epochs, $\eta$ (learn. rate)   		& 2, 0.0003		 	& 4, 0.0003      \\
    $B_i, i>1$: epochs, $\eta$ (learn. rate)   	& 2, 0.0003    	& 2, 0.0003  \\
    \toprule
    \multicolumn{3}{c}{\textbf{CWR+}}   \\
    \cmidrule(r){1-3}
    \multicolumn{1}{c}{\textit{Parameters}}    & \multicolumn{1}{c}{\textit{CaffeNet}}    & \multicolumn{1}{c}{\textit{GoogLeNet}} \\
    \midrule
    Head (see App. A) 							& Maximal  			& Maximal     \\
    $B_1$: epochs, $\eta$ (learn. rate)   		& 2, 0.0003		 	& 4, 0.0003      \\
    $B_i, i>1$: epochs, $\eta$ (learn. rate)   	& 2, 0.0003    	& 2, 0.0003  \\
    \toprule
    \multicolumn{3}{c}{\textbf{AR1}}   \\
    \cmidrule(r){1-3}
    \multicolumn{1}{c}{\textit{Parameters}}    & \multicolumn{1}{c}{\textit{CaffeNet}}    & \multicolumn{1}{c}{\textit{GoogLeNet}} \\
    \midrule
    Head (see App. A) 							& Maximal  			& Maximal     \\
    $\xi$ 										& 1e-7  			& 1e-7     \\
    $w_1, w_i (i > 1)$ 							& 0.0015, 0.0015  	& 0.0015, 0.0015     \\
    $max_F$										& 0.001 			& 0.001  \\
    $\lambda$									& 8.0e5				& 8.0e5 \\
    $B_1$: epochs, $\eta$ (learn. rate)   		& 2, 0.0003		 	& 4, 0.0003      \\
    $B_i, i>1$: epochs, $\eta$ (learn. rate)   	& 2, 0.0003    		& 2, 0.0003  \\
    \bottomrule
  \end{longtable}
\FloatBarrier

\newpage
\section{Hyperparameter Values for iCIFAR-100}
\label{sec:hyperparameter_values_for_icifar_100}

  \begin{longtable}{lll}
  	\caption{the hyperparameter values used for CifarNet \cite{Zenke2017} on iCIFAR-100. The selection was performed on run 1, and hyperparameters were then fixed for runs $2, \dots,10$.}\\[-10pt]
    \label{tab:hyper_icifar100}
  	\endfirsthead
	\endhead
    \toprule
    \multicolumn{2}{c}{\textbf{Cumulative}}                   \\
    \cmidrule(r){1-2}
    \multicolumn{1}{c}{\textit{Parameters}}    & \multicolumn{1}{c}{\textit{CifarNet}}\\
    \midrule
    epochs, $\eta$ (learn. rate) 			& 180, 0.005\\
 
    \toprule
    \multicolumn{2}{c}{\textbf{Naive}}   \\
    \cmidrule(r){1-2}
    \multicolumn{1}{c}{\textit{Parameters}}    & \multicolumn{1}{c}{\textit{CifarNet}}\\
    \midrule
    Head (see App. A) 							& Maximal  		\\
    $B_1$: epochs, $\eta$ (learn. rate)   		& 60, 0.001		 \\
    $B_i, i>1$: epochs, $\eta$ (learn. rate)   	& 60, 0.001    	 \\
    \toprule
    \multicolumn{2}{c}{\textbf{LWF}}   \\
    \cmidrule(r){1-2}
    \multicolumn{1}{c}{\textit{Parameters}}    & \multicolumn{1}{c}{\textit{CifarNet}}\\
    \midrule
    Head (see App. A) 							& Expanding \\
    $map$										& [0.5...0.9] $\rightarrow$ [0.45...0.85]\\
    $B_1$: epochs, $\eta$ (learn. rate)   		& 20, 0.001	\\
    $B_i, i>1$: epochs, $\eta$ (learn. rate)   	& 20, 0.001 \\
    \toprule
    \multicolumn{2}{c}{\textbf{EWC}}   \\
    \cmidrule(r){1-2}
    \multicolumn{1}{c}{\textit{Parameters}}    & \multicolumn{1}{c}{\textit{CifarNet}}\\
    \midrule
    Head (see App. A) 							& Maximal\\
    $max_F$										& 0.001 \\
    $\lambda$									& 8.0e7	\\
    $B_1$: epochs, $\eta$ (learn. rate)   		& 60, 0.001 \\
    $B_i, i>1$: epochs, $\eta$ (learn. rate)   	& 25, 0.00002 \\
    \toprule
    \multicolumn{2}{c}{\textbf{SI}}   \\
    \cmidrule(r){1-2}
    \multicolumn{1}{c}{\textit{Parameters}}    & \multicolumn{1}{c}{\textit{CifarNet}} \\
    \midrule
    Head (see App. A) 							& Maximal \\
    $\xi$ 										& 1e-7  \\
    $w_1, w_i (i > 1)$ 							& 0.00001, 0.00175 \\
    $max_F$										& 0.001 \\
    $\lambda$									& 6.0e7	\\
    $B_1$: epochs, $\eta$ (learn. rate)   		& 60, 0.0005	\\
    $B_i, i>1$: epochs, $\eta$ (learn. rate)   	& 60, 0.00002  \\
    \toprule
    \multicolumn{2}{c}{\textbf{CWR}}   \\
    \cmidrule(r){1-2}
    \multicolumn{1}{c}{\textit{Parameters}}    & \multicolumn{1}{c}{\textit{CifarNet}} \\
    \midrule
    Head (see App. A) 							& Maximal  \\
    $w_1, w_i (i > 1)$ 							& 1, 1  \\
    $B_1$: epochs, $\eta$ (learn. rate)   		& 60, 0.001	\\
    $B_i, i>1$: epochs, $\eta$ (learn. rate)   	& 60, 0.001 \\
    \toprule
    \multicolumn{2}{c}{\textbf{CWR+}}   \\
    \cmidrule(r){1-2}
    \multicolumn{1}{c}{\textit{Parameters}}    & \multicolumn{1}{c}{\textit{CifarNet}} \\
    \midrule
    Head (see App. A) 							& Maximal  \\
    $B_1$: epochs, $\eta$ (learn. rate)   		& 60, 0.001	\\
    $B_i, i>1$: epochs, $\eta$ (learn. rate)   	& 60, 0.001 \\
    \bottomrule
    \pagebreak
    \toprule
    \multicolumn{2}{c}{\textbf{AR1}}   \\
    \cmidrule(r){1-2}
    \multicolumn{1}{c}{\textit{Parameters}}    & \multicolumn{1}{c}{\textit{CifarNet}} \\
    \midrule
    Head (see App. A) 							& Maximal  \\
    $\xi$ 										& 1e-7  \\
    $w_1, w_i (i > 1)$ 							& 0.00015, 0.000005  \\
    $max_F$										& 0.001 \\
    $\lambda$									& 4.0e5	\\
    $B_1$: epochs, $\eta$ (learn. rate)   		& 60, 0.001	\\
    $B_i, i>1$: epochs, $\eta$ (learn. rate)   	& 60, 0.001 \\
    \bottomrule
  \end{longtable}

\section{On Initializing Output Weight to Zero}
\label{sec:on_initializing_output_weight_to_zero}

It is well known that neural network weights cannot be initialized to 0, because this would cause intermediate neuron activations to be 0, thus nullifying backpropagation effects. While this is certainly true for intermediate level weights, it is not the case for the output level.

More formally, let $\theta_{ab}$ be a weight of level $l$, connecting neuron $a$ at level $l-1$ with neuron $b$ at level $l$ and let $net_x$ and $out_x$ be the activation of neuron $x$ before and after the application of the activation function, respectively; then, the gradient descent weight update is proportional to: 

\begin{equation}
	\frac{\partial L_{cross}(\hat{y}, t)}{\partial \theta_{ab}} = \frac{\partial L_{cross}(\hat{y}, t)}{\partial net_b} \cdot \frac{\partial net_b}{\partial \theta{ab}} = \frac{\partial L_{cross}(\hat{y},t)}{\partial net_b} \cdot out_a
  \label{eq:deriv}
\end{equation} 

It is well evident that if $out_a$ is 0, weight update cannot take place; therefore weights of levels up to $l-1$ cannot be all initialized to 0. For the last level (i.e., $l$ coincides with output level), in case of softmax activation and cross-entropy loss, eq. \ref{eq:deriv} becomes \cite{Sadowski}:

\begin{equation}
  \frac{\partial L_{cross}(\hat{y}, t)}{\partial \theta_{ab}} = (\hat{y}_b - t_b) \cdot out_a = (out_a - t_b) \cdot out_a
\end{equation}

and initializing $\theta_{ab}$ to 0 does not prevent the weight update to take place.

\end{document}